\documentclass[10pt]{article} % For LaTeX2e

\usepackage[preprint]{rlj} % Should be uncommented for preprint versions

\usepackage{amsmath,amsfonts,amssymb}            % Defines common symbols like \mathbb R
\usepackage{mathtools}          % Extends amsmath, providing common math tools
\usepackage{mathrsfs}           % Enables \mathscr, which can work in cases that \mathcal does not
\usepackage{graphicx}           % For including images
\usepackage{subcaption}         % Allows for the use of subfigures and subcaptions
\usepackage[space]{grffile}     % For spaces in image names
\usepackage{url}                % For displaying URLs
\usepackage{lipsum}             % For placeholder text
\usepackage{tabularray}
\usepackage{booktabs}
\usepackage{array}
\newcolumntype{H}{>{\setbox0=\hbox\bgroup}c<{\egroup}@{}}
\newcommand{\psim}{p_\textsuperscript{sim}}
\newcommand{\gsim}{g_\textsuperscript{sim}}
\usepackage{algorithm}
\usepackage{algpseudocode}
\usepackage{bbm}

\title{A Meta Reinforcement Learning Approach to Goals-Based Wealth Management}

\setrunningtitle{MetaRL for Goals Based Wealth Management}

\author{Sanjiv R. Das, Harshad Khadilkar, Sukrit Mittal, Daniel Ostrov, Deep Srivastav, Hungjen Wang}

\emails{srdas@scu.edu$^{1}$, harshadk@iitb.ac.in$^{2,7}$, sukrit.mittal@franklintempleton.com$^{3}$, dostrov@scu.edu$^{4}$, deepratna.srivastav@franklintempleton.com$^{5}$, jenrealms@gmail.com$^{6}$}

\affiliations{
$^{1}$\textbf{Dept. of Finance, Santa Clara University, Santa Clara, CA, USA }\\
$^{2}$\textbf{Indian Institute of Technology, Mumbai, India}\\
$^{3}$\textbf{Franklin Templeton, Hyderabad, India}\\
$^{4}$\textbf{Dept. of Mathematics, Santa Clara University, Santa Clara, CA, USA}\\
$^{5}$\textbf{Franklin Templeton, San Ramon, CA, USA}\\
$^{6}$\textbf{Independent, New York, NY, USA}\\
$^{7}$\textbf{Franklin Templeton, Mumbai, India}\\
\par % If including additional comments like below, use \par to add some whitespace. 
}

\contribution{
    \textbf{Generalization}: The MetaRL algorithm covers a remarkably wide set of investment and financial goal scenarios, including different wealths, time frames, goals, and planned infusions, as well as economic regime changes over time during which the expected returns and volatilities of the available investments can shift significantly.
    }
    {
    None
    }

\contribution{
    {\bf Speed:} Because we can use RL inference with the pre-trained MetaRL model, pure inference can determine both what current goals to fulfill or forgo and the best investment portfolio for the next time step over 100 times faster than Dynamic Programming (DP) can. 
    }
    {
    Determining some other properties of the solution, such as the expected utility or the probability of attaining future goals, requires repeatedly querying the MetaRL model, leading to essentially the same runtime as DP for simple cases, but progressively better runtimes compared to DP as the problems become more complicated.
    }

\contribution{
    {\bf Accuracy:} Over 66 new test scenarios, many of which are well outside the training distribution, RL inference achieves (on average) 97.8\% of the expected utility from fulfilled goals compared to the exact optimal solution, which can be obtained by DP for these examples. 
    }
    {
    None
    }
    
\contribution{
    {\bf Extendibility:} Perhaps more important than all of the above points is the fact that the MetaRL algorithm is extendable to more complex GBWM problems that DP cannot address. For example, we consider the GBWM problem in the context of stochastic inflation. For DP, this means changing the state space from two variables to four variables. Due to the so-called ``curse of dimensionality,'' the computational time for DP increases exponentially with the number of state space variables, making this computationally infeasible. In contrast, the two extra state variables don't even noticeably slow down the MetaRL algorithm.
    }
    {
    Of course, this requires re-training the MetaRL model.
    }

 % Your keywords

\summary{Applying concepts related to zero-shot meta-learning and pre-training of foundation models, we develop a meta reinforcement learning approach (denoted MetaRL) that is pre-trained on thousands of goals-based wealth management (GBWM) problems. Each GBWM problem involves a multiple year scenario over which the investor looks to optimally choose an investment portfolio each year and choose to fulfill all, some, or none of the different financial goals that arise each year.  These choices seek to maximize the expected total investor utility obtained from the fulfilled financial goals. By eliminating separate training and optimization for each new investor problem, the MetaRL model in inference mode produces near-optimal dynamic investment portfolio and goal fulfilling strategies for a new GBWM problem within a few hundredths of a second. This delivers expected utilities that are, on average, 97.8\% of the optimal expected utilities (determined via Dynamic Programming). These results are remarkably robust to capital market regime changes, even when training uses only one capital market regime. Further, the MetaRL approach can enable solving problems with larger state spaces where Dynamic Programming becomes computationally infeasible.
}

\begin{document}

\makeCover  % Create the cover page
\maketitle  % Make the title section

\begin{abstract}
Applying concepts related to zero-shot meta-learning and pre-training of foundation models, we develop a meta reinforcement learning approach (denoted MetaRL) that is pre-trained on thousands of goals-based wealth management (GBWM) problems. Each GBWM problem involves a multiple year scenario over which the investor looks to optimally choose an investment portfolio each year and choose to fulfill all, some, or none of the different financial goals that arise each year.  These choices seek to maximize the expected total investor utility obtained from the fulfilled financial goals. By eliminating separate training and optimization for each new investor problem, the MetaRL model in inference mode produces near-optimal dynamic investment portfolio and goal-fulfilling strategies for a new GBWM problem within a few hundredths of a second. This delivers expected utilities that are, on average, 97.8\% of the optimal expected utilities (determined via Dynamic Programming). These results are remarkably robust to capital market regime changes, even when training uses only one capital market regime. Further, the MetaRL approach can enable solving problems with larger state spaces where Dynamic Programming becomes computationally infeasible.
\end{abstract}

%%%%%%%%%%%%%%%%%%%%%%%%%%%%%%%%%%%%%%%%%%%%%%%%%%%%%%%%%%%%%%%%
%% Section: Submission of papers to RLJ/RLC
%%%%%%%%%%%%%%%%%%%%%%%%%%%%%%%%%%%%%%%%%%%%%%%%%%%%%%%%%%%%%%%%
\newpage

\section{Introduction}

Goals-based wealth management (GBWM) is an investment paradigm centered on maximizing the attainment of multiple financial goals over time, weighted by their relative priority. To achieve this, an investor must continually optimize two interconnected choices in managing their wealth over time: (1) they must determine which goals to pursue and which to abandon to preserve the capital required for high-priority future aspirations, and (2) they must dynamically adjust their investment portfolio in response to both evolving market conditions and progress toward specific objectives. In contrast, traditional wealth management generally neglects these important choices, as well as individual nuance, instead assigning investors into one of three to five broad risk-appetite brackets, each with a corresponding static ``one-size-fits-all'' investment portfolio. This reductive approach frequently leaves investors' specific financial needs unmet.

This paper introduces a novel {\it meta-model} approach that leverages the power of artificial intelligence and deep neural networks to determine an individualized, optimized combined goal-attainment and dynamic investment strategy. Our approach addresses the GBWM problem without requiring an explicit, case-by-case solution. We train a reinforcement learning (RL) model across thousands of heterogeneous GBWM scenarios to optimize financial outcomes. We demonstrate that this pre-trained meta-model, designated  ``MetaRL,'' can be queried via inference to produce near-optimal strategies for new financial problems in mere hundredths of a second. To our knowledge, such a meta-model has not been established in the finance literature for GBWM or analogous optimization problems. While the current state of the art in GBWM mostly employs dynamic programming (DP) (see \cite{bellman_theory_1952, bellman1966dynamic, das_dynamic_2022, capponi_continuous_2023}), MetaRL offers significant computational and structural advantages over DP, as detailed in Subsection \ref{MvsDP}.

The MetaRL architecture mirrors the logic of large language models (LLMs), which are {\it pre-trained} on massive text corpora (e.g., \cite{brown_language_2020, wei_finetuned_2022, gemma_team_gemma_2024}) to perform diverse tasks without further training. Similarly, MetaRL is trained on thousands of GBWM instances encompassing a wide array of financial circumstances. Analogous to an LLM in inference mode, MetaRL provides {\it zero-shot} solutions for both portfolio selection and goal-taking decisions. Our methodology is further inspired by meta-learning frameworks such as \cite{finn_model-agnostic_2017}, which utilize meta-models to navigate tasks with previously unseen environmental parameters.

We adopt a unique set of normalizations of the state variables of the problem, which enables the MetaRL algorithm to be used in inference mode and remain remarkably robust. Under this proposed MetaRL framework, we can seamlessly modify parameters such as (i) investor goals (quantity, timing, cost, and priority), (ii) capital constraints such as time horizon, initial wealth, and the frequency/magnitude of future infusions, and (iii) the investable universe, i.e., the number of available portfolios and their positions on the efficient frontier (the Pareto optimal curve of expected returns versus volatility, described in Section \ref{EFront}).
Notably, we can even allow the efficient frontier itself to fluctuate, accounting for the inherent uncertainty of capital markets. In contrast, if we use dynamic programming, a change to even one of these variables necessitates re-solving the entire optimization problem from scratch. 

We implement a two-agent variant of Proximal Policy Optimization (PPO) (see \cite{schulman_trust_2015, schulman_proximal_2017, haarnoja_soft_2018}) to construct the MetaRL model. The core methodology is outlined in Section \ref{sec:dprl}, with exhaustive technical specifications provided in Appendices \ref{RLalgo} and \ref{sec:training_inference}. Our results indicate that for a given parameter set, RL inference (utilizing 26 state variables) is, on average, over 100 times faster than DP (utilizing two state variables) at identifying the optimal current strategy. Furthermore, the MetaRL model yields approximately 97.8\% of the accrued utility produced by the exact optimal strategy derived from DP.

While recent DP methodologies \citep{das_dynamic_2022, capponi_continuous_2023} significantly outperform fixed-rule or Monte-Carlo strategies, they lack the scalability required for high-throughput applications. The MetaRL model presented here utilizes deep neural network policy functions that map state variables directly to policy actions. By implementing a dual-PPO algorithm, we decouple investment portfolio selection and goal-taking decisions into two distinct actor networks. This work complements existing meta-learning research \citep{gupta_meta-reinforcement_2018, finn_meta-learning_2018, beck_survey_2023} by offering a model that is generalizable to unseen economic scenarios, computationally superior to DP, and extensible to higher-order stochastic effects like inflation (see Subsection \ref{sec:stochastic_inflation}).

The paper proceeds as follows. In Section \ref{sec:gbwm_technical}, we explain the GBWM problem under the assumption that the investor has, at most, one all-or-nothing goal each year. In Section \ref{sec:dprl}, we give a brief overview of the features of the MetaRL approach we use to solve this problem, relegating most of the technical details about the MetaRL model to Appendices \ref{RLalgo} and \ref{sec:training_inference}. In Section \ref{sec:results} we explore experimental results and examples showing the runtimes, the produced strategies, the accuracy of the MetaRL approach, and its robustness to changes to the available investment portfolios. Extensions of the model are taken up in Section \ref{sec:extensions}, where concurrent and partial goals are introduced in Subsection \ref{subsec:concurrent_partial_goals}, while Subsection \ref{sec:stochastic_inflation} extends the GBWM problem to include the effect of stochastic inflation, illustrating how the algorithm is easily extended to handle more state variables. We offer a detailed discussion of the related literature in Section  \ref{sec:litdetailed}, which is best read with full context after reading this paper. A concluding discussion is provided in Section \ref{sec:conclusion}.

\section{Defining The GBWM Problem}
\label{sec:gbwm_technical}

\subsection{Decisions Facing Goals-Based Investors}

An investor using goals-based wealth management (GBWM) wishes to attain as many of their financial goals, weighted by their importance to the investor, as possible over the time horizon of their portfolio, which may mean throughout their projected lifetime. To weight the importance of each goal to an investor, a utility (reward) is assigned to each goal. Realistic methods for determining these utilities come from the investor balancing the probabilities of attaining their goals against each other, as discussed in \citet[]{das_dynamic_2022} and \citet[]{das_efficient_2023}. The investor is then looking to make decisions that will optimize their ``expected attained utility,'' meaning the expected value of the utilities of attained goals in the present and future.

To optimize the expected attained utility, there are two decisions that must be optimized at each time step within the time horizon: (1) {\it Goal-taking}: whether or not to pay the cost needed to fulfill a currently available goal and attain the utility assigned to that goal, and (2) {\it Portfolio choice}: determining which investment portfolio (level of expected return and its corresponding volatility) to hold for the next time step. Because we are in a dynamic setting, where the optimization must be in light of knowing that these decisions can be different in every future time step, this is a complicated optimization problem. Fulfilling a goal requires there to be sufficient wealth available, but even when there is, the investor may be better off forgoing the current goal to increase their probability of fulfilling more important future goals. We note that goals cannot be deferred to later time steps in our model. They are either fulfilled or not at the times when they are scheduled to occur.

The investment portfolio decision also depends on how much wealth the investor has. If the portfolio is not doing well, the investor may need to select more aggressive investment portfolios in a measured way to increase the probability that they can achieve more future goals. On the other hand, if the portfolio is doing very well, the investor may wish to move to more conservative investment portfolios to avoid large losses that could adversely impact their ability to fulfill future goals. 

% The investor's problem may be solved using techniques such as Dynamic Programming (DP) and reinforcement learning (RL), which we will compare. This paper introduces a large-scale extension of RL, which we denote as MetaRL. 

\subsection{Parameters Defining A Single Problem}
\label{sec:param}

Each investor has their own GBWM problem, which must be defined so that the GBWM solution can be optimized for their individual circumstances. The parameters that must be specified to define a single investor GBWM problem are the following:

\begin{enumerate}
    \item {\it Time horizon and time steps:} The investor must specify $T$, the total number of time steps, and $h$, the amount of time between time steps. We use $t=0,1,2,...,T$ to denote the time step's index. Further, because $h$ is fixed to equal one year in all of this paper's experiments, going forward we will refer to $t$ as the time (in years) and $T$ as the time horizon (in years) of the portfolio.
    \item {\it Initial wealth and future wealth infusions:} The investor must specify their initial wealth, $W(0)$, where $W(t)$ denotes their wealth at time $t$. Wealth infusions can be specified at any time steps. The amount of money in an infusion at time step $t$ is denoted as $I(t)$ (where $I(t)=0$ if there is no infusion in year $t$), or, collectively, as the $T$-vector ${\bf I}$. 
    % We will, for the moment, assume that ${\bf I}={\bf 0}$ (no infusions), and then consider ${\bf I}\neq{\bf 0}$ in Section \ref{Infu}.
    \item {\it Goals:} For simplicity, we will, for the moment, assume there can be at most one all-or-nothing goal available at each time step. (We will extend this to allowing concurrent and partial goals in Subsection \ref{subsec:concurrent_partial_goals}, which will require extending the MetaRL model.) With this assumption, the investor can specify $C(t)$, the cost of the goal available in year $t$ (where $C(t)=0$ if there is no goal available in year $t$), or, collectively, as the $T$-vector ${\bf C}$. The corresponding utilities are then specified by $U(t)$ (where, again, $U(t)=0$ if there is no goal in year $t$), or, collectively, as the $T$-vector ${\bf U}$. Note that an investor would be just as happy attaining one goal with a utility of 4 and another goal with a utility of 7 as they would be attaining just one goal with a utility of 11. 
    \item {\it Available investments:} The investor (or, more likely, the financial services company working with the investor) must also specify the expected return and the volatility of each of the $P$ available investment portfolios to the investor, which we denote respectively by $\mu_p$ and $\sigma_p$, where $p\in\{0,1,\ldots,P-1\}$, or, collectively, by the $P$-vectors $\boldsymbol{\mu}$ and $\boldsymbol{\sigma}$. The investment portfolio choice, $p$, can be changed in each time step.
\end{enumerate}

%These parameters that define a single scenario are summarized in Table \ref{tab:stateinputs}.

% \begin{table}[h!]
% \centering
% \caption{Scenario parameter variables: This list of variables constitutes a complete description of the parameters that define a single scenario.}
% \label{tab:stateinputs}
% {\small
% \begin{tabular}{|p{1.5cm}|p{2.4cm}|p{5.8cm}|}
% \hline
% Symbol & Length (And Type) & Explanation \\ \hline
% $T$ & 1 (int) & Number of time steps (also called the ``time horizon'') \\
% $h$ & 1 (float) & Time between time steps (assumed to be a year throughout this paper) \\
% $W(0)$ & 1 (float) & Initial wealth at time $t=0$ \\
% ${\bf{I}} $ & $T$ (float) & Planned infusions of wealth at each time step where $t>0$\\
% ${\bf{U}}$ & $T$ (float) & Utility of each time step's goal where $t>0$ \\
% ${\bf{C}}$ & $T$ (float) & Cost of each time step's goal where $t>0$ \\
% $\boldsymbol{\mu}$ & $P$ (float) & Expected return of each of the $P$ investment portfolios \\
% $\boldsymbol{\sigma}$ & $P$ (float) & Volatility of each of the $P$ investment portfolios \\
% \hline
% \end{tabular}
% }
% \end{table}

\subsection{Mathematical Formulation Of The Problem}

The two decisions the investor must make at each time step are defined mathematically by 
\begin{enumerate}
    \item {\it The goal-taking decision:} At each time, $t$, the investor first decides whether or not to take the goal if one is available. We denote this decision by $g(t) \in \{0,1\}$, where $g=0$ means the investor decides not to take the goal (or can't afford it) and $g=1$ means the investor decides to take the goal.
    \item {\it The investment portfolio decision:} After the goal-taking decision is made, so the investor knows how much of their wealth remains, they decide which investment portfolio to select for the next time step. We denote their choice by $p(t) \in \{0,1,...,P-1\}$, where $\mu_{p(t)}$ and $\sigma_{p(t)}$ are the expected return and volatility of this investment portfolio choice.
\end{enumerate}
With the parameter variables of the problem from Section \ref{sec:param} and the two decision variables defined, we can mathematically define the goal of the GBWM problem, which is to optimize the expected utility  attained (see also Appendix \ref{RLalgo}); that is, we seek to attain 
\begin{equation}
\max_{g(t),\, p(t),\, {\mbox{\scriptsize for }} t=\{0,...,T-1\}} E\left[\sum_{t} g(t)\cdot U(t) \right],  \label{gbwm_obj}
\end{equation}
subject to wealth transitions, which we assume are governed by geometric Brownian motion, meaning
\begin{equation}\label{wealth_transition}
W(t+1) = [W(t)+I(t)-g(t)\cdot C(t)]\cdot \exp \left[ \left(\mu_{p(t)} - \frac{1}{2}\sigma_{p(t)}^2 \right)h + \sigma_{p(t)} \sqrt{h} \cdot Z \right],  
\end{equation}
where $Z \sim N(0,1)$ is a standard normal random variable. While using geometric Brownian motion is a standard assumption for a wealth evolution model in the finance literature, it is not a requirement for our RL formulation, which can accommodate any simulatable stochastic wealth evolution model. For example, we may decide, as is common, to replace the standard normal $Z$ in equation \eqref{wealth_transition} with a Student's $T_n$-distributed random variable where $n$, the degrees of freedom, can be any desired positive integer.  
% More detailed equations for these wealth transitions are given in Appendix
% \ref{sec:env_wealth_transition}. 

\section{Solution: MetaRL}
\label{sec:dprl}

\begin{figure}
    \centering
    \includegraphics[width=0.95\linewidth]{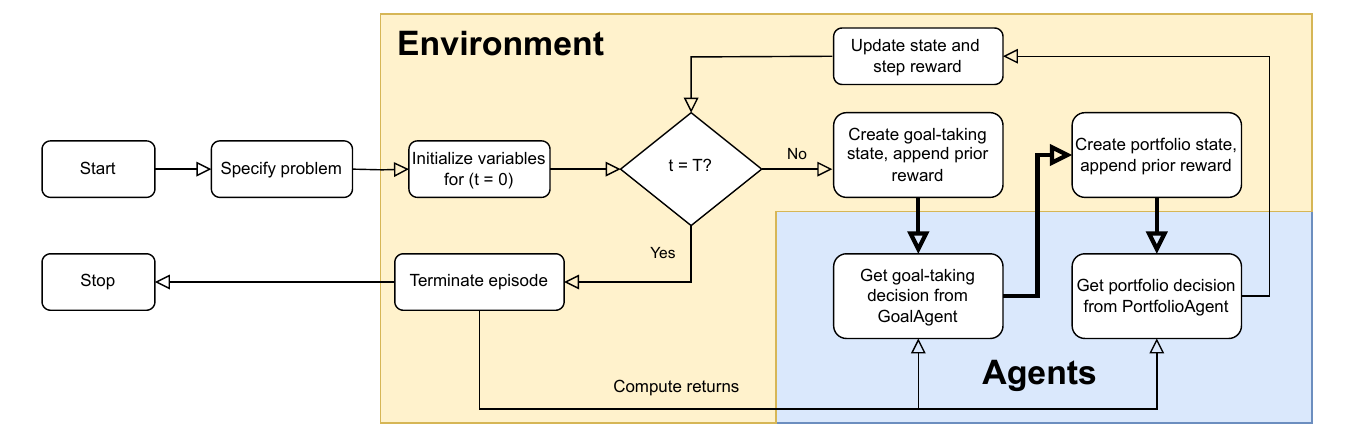}
    \caption{Logical flowchart of the MetaRL approach.}
    \label{fig:flowchart}
\end{figure}

We give a brief overview of the states, actions, and rewards used to train the MetaRL model. Details about this and other aspects of the training are available in Appendix \ref{RLalgo} and \ref{sec:training_inference}. A high-level overview of the execution of each episode (in training or testing) is shown in Figure \ref{fig:flowchart}. While the agent and environment are largely standard, the novel aspect is the existence of two agents: one for goal-taking, the next for portfolio selection. Note that the goal-taking agent has a causal effect on the observations of the portfolio agent. The rewards are designed accordingly, as described below.

\subsection{State Space}\label{StSp}

At first thought, we might consider choosing the state space variables to simply be the problem parameter variables given in Subsection \ref{sec:param}. However, this direct form of the variables is not particularly helpful for MetaRL, since there are inter-relationships between the variables (for example, goal costs and portfolio returns) which can be difficult to learn. Furthermore, handling an arbitrary number of portfolio options and time horizons requires a scale-invariant state space definition. Our state variables are designed to be largely dimensionless, helping with generalizability. We next give a brief overview of the 26 state variables used by both agents (GoalAgent and PortfolioAgent). These are explained in far more detail in Appendix \ref{sec:state}. 

The first state variable is the current time, $t$, normalized (that is, divided) by $T$, the time horizon. The second and third state variables are the current (i.e., time $t$) wealth, $W(t)$, normalized by the costs of current and future goals discounted to time $t$ dollars using a pessimistic investment return projection for one state variable and an optimistic investment return projection for the other state variable.

The next three sets of variables describe the utilities and costs of current and future goals. The first set is, essentially a vector of the utilities in each current and future year normalized by the sum of these utilities. Because the length of this vector would vary with time $t$ and we need a vector of constant length for training, we instead clump these results into time blocks. In our case, we choose seven time blocks corresponding to 0, 1, 2, 3, 4--5, 6--9, and 10 or more years into the future. So, the 6--9 time block contains the sum of the utilities for goals that are between six and nine years into the future from time $t$ divided, as before, by the sum of the utilities of all current and future goals. For the costs, we do the same thing, but we again use the costs discounted to time $t$ dollars both using a pessimistic and an optimistic investment return projection. This gives seven normalized state variables for the utilities and $2\times7=14$ normalized state variables for the costs.

The remaining two state variables are closely connected to the two decisions, and we think of them as ``indicator states.'' The first, connected to goal-taking, gives a very rough estimate for comparing the effect on the overall attained utility if the investor takes the current (time $t$) goal versus if they do not take it. The computation is carried out by a deterministic Monte-Carlo simulation of both possible actions, with a fixed set of representative return values for each portfolio. This variable is normalized between 0 and 1, where the closer it is to 1, the more evidence there is to take the goal. The second variable, connected to selecting the investment portfolio, runs a similar deterministic Monte-Carlo simulation with a constant portfolio choice up to the end of tenure, to indicate the portfolio which returns the highest expected utility.

\subsection{Actions And Decisions}\label{MAD}

Appendix \ref{sec:env_wealth_transition} shows how the MetaRL algorithm uses the 26 state variables from the previous subsection to produce two continuous {\it actions}, $a_g(t) \in [0,1]$ and $a_p(t)\in [0,1]$, that directly determine the discrete {\it decisions} $g(t) \in \{0,1\}$ and $p(t) \in \{0,1,...,P-1\}$ via simple, monotonic mappings given in equations \eqref{gdef} and \eqref{pdef} of Appendix \ref{sec:env_wealth_transition}. There are three reasons the MetaRL actions are chosen to be continuous variables instead of discrete: (1) If $a_p(t)$ were discrete, we would need to set $a_p(t) = p(t)$, which would mean needing to re-train the MetaRL model each time $P$ changed, which is undesirable. (2) If $a_p(t)$ were a discrete action, there would be no relationship between the investment portfolios from the MetaRL perspective, which would lose the important connection that the investment portfolios become more aggressive as $p$ increases. (3) If $a_g(t)$ were a discrete action, we would be unable to infer the confidence with which the action was being chosen, which would affect the interpretability of the model.

\subsection{Rewards}

The expected attained utility given in equation \eqref{gbwm_obj} is the extrinsic reward that the MetaRL model needs to optimize. To this we add a dense intrinsic reward signal based on the discrepancy between the indicator states explained above (for goal-taking and portfolio selection respectively) and the action outputs $a_g(t)$ and $a_p(t)$, as explained in Appendix \ref{sec:Rew}. The intrinsic reward is annealed to a low value over the course of training, in order to emphasize the true objective.

\subsection{Training Regime}\label{TR}

The training of the MetaRL approach depicted in Figure \ref{fig:flowchart} is carried out with five random seeds, with each seed trained on 1000 randomly generated examples, as detailed in Appendix \ref{sec:curriculum}. The training results are shown in Figure \ref{fig:training}, which is also in Appendix \ref{sec:curriculum}. Intuitively, the synthetic training scenarios were designed to thoroughly explore the normalized state space defined in Table \ref{tab:observations} by varying key drivers of fundedness and uncertainty. Initial wealth and infusions were scaled relative to goal costs to vary between the extremes of under-fundedness and over-fundedness, while the number of goals and investment horizons were adjusted to capture different levels of economic uncertainty. Training horizons ranged from 5 to 50 years. Although a single efficient frontier was used in training, varying horizons effectively introduced uncertainty comparable to changing financial regimes, as we will see in Section \ref{EFront}.
In the results that follow, all actions are computed by running inference on all five trained models (from the five random seeds) and taking a statistical measure, such as a fixed percentile like the median, over all five actions.

\subsection{Comparison with Dynamic Programming}\label{MvsDP}

Dynamic Programming (DP) is a highly relevant algorithm for solving the optimization problem \eqref{gbwm_obj}, since it allows us to directly optimize for utility using the Bellman operator (see \cite{bellman_theory_1952}). For any given problem, DP uses a backwards pass, meaning a computation that starts at the final time $t=T$ and works backwards to the starting time $t=0$, to determine the optimal decisions and the optimal ``DP value function,'' meaning the optimal expected attained utility from current and future goals. The backwards pass computes the optimal decisions and the optimal DP value function at any state, meaning any given time, $t$, and wealth at that time, $W(t)$. We note that the state space in DP contains just these two state variables: $t$ and $W(t)$. However, DP faces the following challenges:
\begin{enumerate}
    \item It has to solve each problem from scratch, whereas MetaRL is able to leverage pre-training.
    \item It requires discretization of both the state and the action space.
    \item Computational time blows up exponentially with the number of state variables due to the ``curse of dimensionality." Realistically, this generally limits DP to three or fewer state variables.
    \item Because of the backward pass nature of DP, it is unable to handle dynamic constraints based on prior actions, such as limiting $|p(t)-p(t-1)|$, the change from the previous investment portfolio.
\end{enumerate}

%MetaRL is different. First, the definition of the ``RL value function'' is different from the ``DP value function'' in two ways: (1) The RL value function uses the MetaRL reward, which adds both the extrinsic reward and the intrinsic reward, whereas the DP value function is based solely on the extrinsic reward, meaning the expected attained utility. (2) The RL value function allows the current action variables to be inputs so their effect on the rewards can be gauged, whereas the DP value function requires the decisions to be set at their optimal values.

%We emphasize a second key difference between MetaRL and DP: MetaRL uses 26 state variables, whereas DP only uses 2. There is a considerable advantage in using 26 state variables: they encompass the scenario parameters from Subsection \ref{sec:param} as well as $t$ and $W(t)$, and this is why the MetaRL model does not have to be re-trained when we change scenarios. The fact that RL can work with 26 (or more) state variables in a reasonable amount of time is its key advantage over DP. We use 2 state variables with DP because we have to. We can't use more due to the  curse of dimensionality on DP's computational time, discussed earlier. This restriction means that DP must be rerun any time we change scenarios, as well as the fact that it cannot be applied to more sophisticated GBWM models that require more than 2 state variables to describe its state, even within a specific scenario (like in Subsection \ref{sec:stochastic_inflation}, which considers the effect of stochastic inflation).

\section{Results} \label{sec:results}

As discussed briefly in Section \ref{TR} and more thoroughly in Appendix \ref{sec:training_inference}, the dual-agent PPO model for MetaRL is trained on 1000 randomly generated problems and 5 random seeds. The resulting trained actor networks, GoalAgent and PortfolioAgent, are used in this section to solve new investor problems. We refer to this solution process as ``RL inference,'' which we will evaluate using specially designed test problems not seen during training. %We first describe (in Subsection \ref{sec:test_suite}) the test suite of 66 GBWM scenarios that we will use in this section. We then explore the speed of using RL inference (in Subsection \ref{sec:CRA}), the goal-taking and portfolio investment decisions determined by using RL inference (in Subsection \ref{sec:DA}), the average attained utility these RL inference determined decisions lead to (in Subsection \ref{sec:EUA}), and the effect on the average attained utility if the efficient frontier changes (in Subsection \ref{EFront}). 
In all cases, we will compare our RL inference results to the optimal decisions that can be generated using DP, noting that DP must be completely rerun for each new problem.

% which is feasible because DP only has two state variables, time $t$ and wealth $W$, to consider. This will not be the case in more sophisticated GBWM problems like Subsection \ref{sec:stochastic_inflation}  where we add the effect of stochastic inflation, which would require DP to use four state variables, making it computationally infeasible.

% {\bf Caveat:} One of the key tests we consider here (Section \ref{EFront}) is determining how robust our results are to changes in the efficient frontier, which correspond to the expected returns and volatilities of our available investment portfolios. Note that two sources of model risk are relevant. First, estimation error in return and covariance parameters effectively alters the frontier; our results indicate that the MetaRL policy is robust to such shifts and can be refreshed through periodic rebalancing when parameters change materially. A second possibility is that there are structural errors in the model, such as departures from the assumed GBM dynamics. While we do not carry out experiments on this source of uncertainty, we note that structural modeling changes do not undermine the RL framework itself. Alternative stochastic processes can be incorporated into the training or inference environment, with little to no change in the training and inference workflows.

\subsection{Test Suite Used For Evaluating The Performance Of The MetaRL Model}
\label{sec:test_suite}

We designed a suite of 66 new investor problems with varying parameters for the time horizon, the initial wealth, the number of goals along with the timing, cost, and utility of each goal, and the number of infusions along with the timing and amount (i.e., worth) of each infusion. A list of these 66 test problems is available in Appendix \ref{66}. These test problems have time horizons that range from 3 to 100 years (with a mean of 38 years), and a range of 1 to 60 goals (with a mean of 16 goals).
We note that the parameters in many of these 66 cases are outside the range of the parameters that were used to train the MetaRL model. For example, a number of our 66 cases use a time horizon of 100 years, whereas the training only uses time horizons up to 50 years.
In our 66 problems and in training the MetaRL model, we consider $P=15$ possible investment portfolios, where $p$ increases as the investment portfolio becomes more aggressive. In our figures, the investment portfolios are labeled from $1$ (the most conservative) to $P=15$ (the most aggressive), which is a more natural numbering created by simply adding 1 to $p \in \{0,1,...,P-1\}$. The 15 investment portfolios used for the problems and for training the MetaRL model range along the baseline efficient frontier discussed in Subsection \ref{EFront}, where we also discuss the fact that the MetaRL model is robust to changing the efficient frontier without requiring re-training.

\subsection{Computational Runtime Analysis: Comparing RL Inference To DP}
\label{sec:CRA}

Solving GBWM problems using DP can be made quite efficient by using techniques like vectorization and just-in-time compilation to C executable code (as has been done for all DP computations in this paper). The RL code has been similarly implemented, with C executable code being especially used to compute the two indicator states described in Subsection \ref{StSp}. Importantly, DP requires computing the entire solution for each problem from scratch, while RL is able to run using pure inference. Here we quantify the resulting computational implications, using the test suite of 66 investor problems described above.
%
%More importantly, as indicated before, as the dimension of the state space (i.e., the number of state space variables) for DP problems grows, DP will quickly become infeasible due to exponential computational time growth, whereas RL inference's computational time does not grow exponentially. In an application where RL models are intended to solve hundreds of problems in a few minutes, such as in a robo-advising setting, the algorithmic and corresponding software improvements in computational speed that RL inference can provide become crucial. In this section, we analyze the computational time needed to run RL inference on the 66 scenarios in our test suite from Subsection \ref{sec:test_suite} using the already obtained MetaRL model versus the computational time needed to run DP on the same test suite. 
%
Two types of computational runtime comparisons may be made: (1) comparing the average time for policy inference at a given time for a given problem, meaning we want to determine the optimal actions/decisions to make at the current time, and (2) comparing the average time it takes to determine the optimized expected attained utility at a given time for a given problem, meaning we want to determine the DP value function at the current time. We will see that using RL inference is over 100 times faster than DP in the first type of comparison, while the computational times for RL inference and DP are comparable in the second type of comparison. %For the remainder of this subsection, we explore these two cases.

%{\color{red}[/HK: Should we point out that all RL results (computation times, heatmaps, average utilities) are produced by running the same computed state inputs through 5 independently trained PPO actor networks (which differ only by the random seed set at the start of training)? The implemented RL actions are the median of the 5 actions returned by the 5 networks.] {\color{blue} [[[DO: Yes, but this should be in Appendix C. Please feel free to change things there to incorporate this!]]]}}

{\it First type (Computational time to determine the optimal actions/decisions at the current time):} To determine the optimal goal-taking decision and the optimal investment portfolio decision at the current time requires DP to run its entire backwards pass to solve for the individual investor's specific problem. For RL inference, on the other hand, the problem does not need to be solved at all, as the MetaRL algorithm is already trained over a representative sample of problems and can be directly used to determine the optimal actions for goal-taking and selecting an investment portfolio at a given time and wealth. %We again note (as discussed in Subsection \ref{MAD}) that these {\it actions} are continuous variables in RL, which are easily (and essentially instantaneously) converted to the final (discrete) goal-taking and investment portfolio {\it decisions} via equations \eqref{gdef} and \eqref{pdef} in Appendix \ref{sec:env_wealth_transition}.

Because RL inference only requires querying the MetaRL model once to obtain these actions, it is far faster than DP, as we can clearly see by looking at the ``mean'' column in Table \ref{tab:inference_runtimes_comparison} (noting that results in the table are in milliseconds (msec)). At times when there is no goal available, so only an investment portfolio choice must be made, RL inference is approximately twice as fast as when both a goal-taking choice and an investment portfolio choice must be made. This can be seen by comparing the mean of 9.277 msec to the mean of 20.94 msec in the table. DP takes an equal amount of computational time either way, but we note that, approximately on average, it is $2198/20.94 = 105.0$ times longer than RL inference at times with a goal and $2198/9.277 =236.9$ times longer than RL inference at times without a goal. That is, even for this GBWM problem, where there are only two state variables used in Dynamic Programming and 26 state variables used by RL inference with the MetaRL approach, we find that RL inference produces optimized decisions more than 100 times faster than DP can.

\begin{table}[!h]
\centering
\caption{\label{tab:inference_runtimes_comparison} %\small
Average time (in milliseconds) to determine the current optimal actions/decisions over the 66 test suite cases, or, for the second row, the 48 of these 66 cases that have more than one goal. 
% See the text for a more thorough explanation of the calculations.
}
%{\footnotesize
\begin{tabular}{p{1.3in}|r|r|r|rrrrr}
\toprule
 & \# of cases & mean & std.~dev. & min & 25\% & 50\% & 75\% & max \\
\midrule
Time (in msec) using RL inference to determine optimal portfolio actions only & 66 & 9.277 & 2.76 & 3.80 & 8.95 & 9.76 & 10.49 & 18.40 \\
Time (in msec) using RL inference to determine optimal portfolio and goal actions & 48 & 20.94 & 1.34 & 16.5 & 20.2 & 21.1 & 21.6 & 23.9 \\
DP backward pass comp. time (in msec) & 66 & 2198 & 2255 & 83 & 579 & 1248 & 3720 & 8012 \\ 
\bottomrule
\end{tabular}
%}
\end{table}

%To more precisely explain the calculations presented in Table \ref{tab:inference_runtimes_comparison}, we describe in more detail how, as an example, we computed the ``50\%'' column, which gives the median results: (1) The first row's result means half of the 66 cases require an average of 0.00976 seconds or less to run RL inference to estimate the investment portfolio decision, where this average is over all times within the case where there is no goal. (2) The second row's result means half of the 48 cases in the 66 test suite examples that have at least one goal before the final time step require an average of 0.0211 seconds or less to run RL inference to estimate both the goal-taking and investment portfolio decisions, where the average is over all times$<T$ within the case where there is a goal. (3) The third row's result means that half of the 66 cases require an average of 1.248 seconds or less to run DP's backwards pass, which determines the optimal goal-taking and investment portfolio decisions, where the average is over all times ($0,1,...,T-1$) within the case. 

{\it Second type (Computational time to determine the optimal expected attained utility at the current time):} The backwards pass that DP just used to determine the optimal goal-taking and investment portfolio decisions also yields the optimal expected attained utility (i.e., DP's value function), so the time DP takes for this second type of computation is still given in the last row of Table \ref{tab:inference_runtimes_comparison}. On the other hand, using RL inference to find the optimal expected utility is more complicated. As we discuss in more detail in the next subsection, we repeatedly query the MetaRL algorithm to determine the optimal decisions over a dense grid in the $(t,W)$ state space plane. We use these decision graphs to specify decisions throughout the evolution of each of 10,000 simulated Monte Carlo trajectories, and then average these 10,000 attained utilities to estimate the optimized expected attained utility. This is necessary because the MetaRL algorithm only provides actions, not DP value function estimates.\footnote{Readers might wonder if the critic network can provide the estimate of DP's value function, however, a broad range of empirical studies of actor-critic architectures have found that the values learned by the critic are indicative rather than exact. For example, see Section III-D of \url{https://hal.science/hal-00756747/document} or page 2 of \url{https://arxiv.org/pdf/1910.08412}.}  

\begin{figure}
\centering
\includegraphics[width = 2.05in, height = 2.05in]{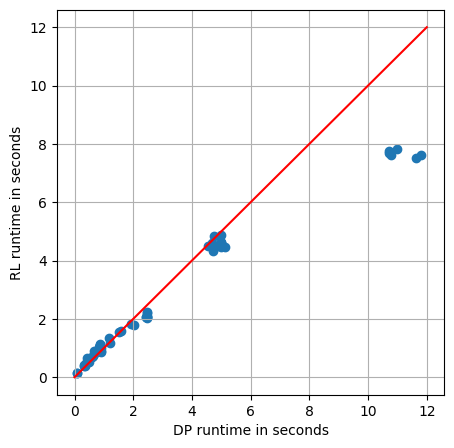}
\hspace{3 em}
\includegraphics[scale=0.5]
{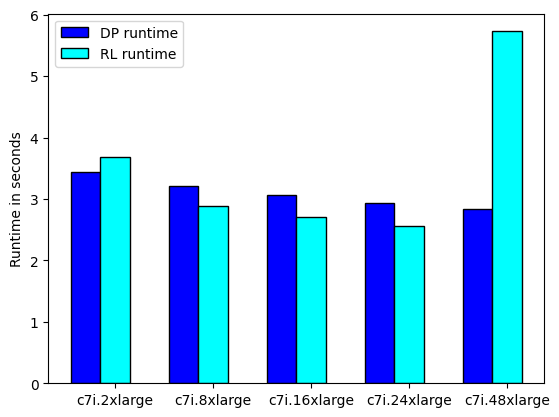}
\caption{\label{fig:dp_rl_runtime} \small Left panel: Plot of each of the 66 test suite GBWM problems comparing their runtimes (in seconds, with the c7i.24xlarge machine) using DP's backwards pass versus using RL inference (and 10,000 Monte Carlo simulations). Right panel: Comparison of average runtimes (in seconds) over the 66 test suite GBWM scenarios for DP and RL inference on various hardware. Compute-optimized machine instances on AWS were used for these experiments. (The prefix `c' in the `c7i.*' stands for compute-optimized.) 
The details of these machines are as follows: (i) c7i.2xlarge -- 8 CPUs (1 socket, 4 cores, 2 threads, 32GB RAM); (ii) c7i.8xlarge -- 32 CPUs (1 socket, 16 cores, 2 threads, 64GB RAM); (iii) c7i.16xlarge -- 64 CPUs (1 socket, 32 cores, 2 threads, 128GB RAM); (iv) c7i.24xlarge -- 96 CPUs (1 socket, 48 cores, 2 threads, 192GB RAM); (v) c7i.48xlarge -- 192 CPUs (2 sockets, 48 cores, 2 threads, 384GB RAM).}
\end{figure}

This leads to DP and RL inference having comparable runtimes for calculating the optimal expected attained utility starting at the current time.  This can be seen in the left panel of Figure \ref{fig:dp_rl_runtime}, where a comparison of the runtimes for each of our 66 test problems is shown. Overall, we see that RL inference runtimes are quite fast, averaging a little better than the 2.198 seconds DP requires, on average, for our test suite of 66 problems. Since the points that are the farthest to the right on the plot correspond to larger problems (longer horizons, more goals) that require greater runtimes, we see from Figure \ref{fig:dp_rl_runtime} that RL inference becomes progressively better than DP in terms of runtime as the problems become larger.

In addition to analyzing algorithms and their software implementation, it is important to assess the role of hardware on runtimes. The right panel in Figure \ref{fig:dp_rl_runtime} shows how the choice of server and its configuration matters. This plot clearly shows that DP and RL inference have comparable runtimes. The main difference between the two comes from the number of CPU cores. At first, as the number of CPU cores increases, RL inference's runtime becomes lower than DP's. However, simply choosing a large machine does not always mean a better runtime, as can be seen when the {\tt c7i.48xlarge} machine is chosen and RL's runtime rises sharply. This sudden rise is because the {\tt c7i.48xlarge} machine has two sockets (split chip), and the cross-chip traffic slows down RL inference. We therefore have run all our problems in this paper (both DP and RL inference) on the {\tt c7i.24xlarge} machines, since it gives the best combination of DP and RL inference runtimes.

 %This can also be seen in Table \ref{tab:inference_runtimes_comparison} by comparing the max and 75\% columns to the median, and noting that the significant increase seen in the DP computational times is not reflected in the RL inference computational times. 

\subsection{Decision Analysis: Comparing RL Inference To DP}\label{sec:DA}

\begin{figure}
\centering
DP: Investment Portfolio Decisions \hspace{.55in}RL Inference: Investment Portfolio Decisions\\
\includegraphics[scale=0.38]{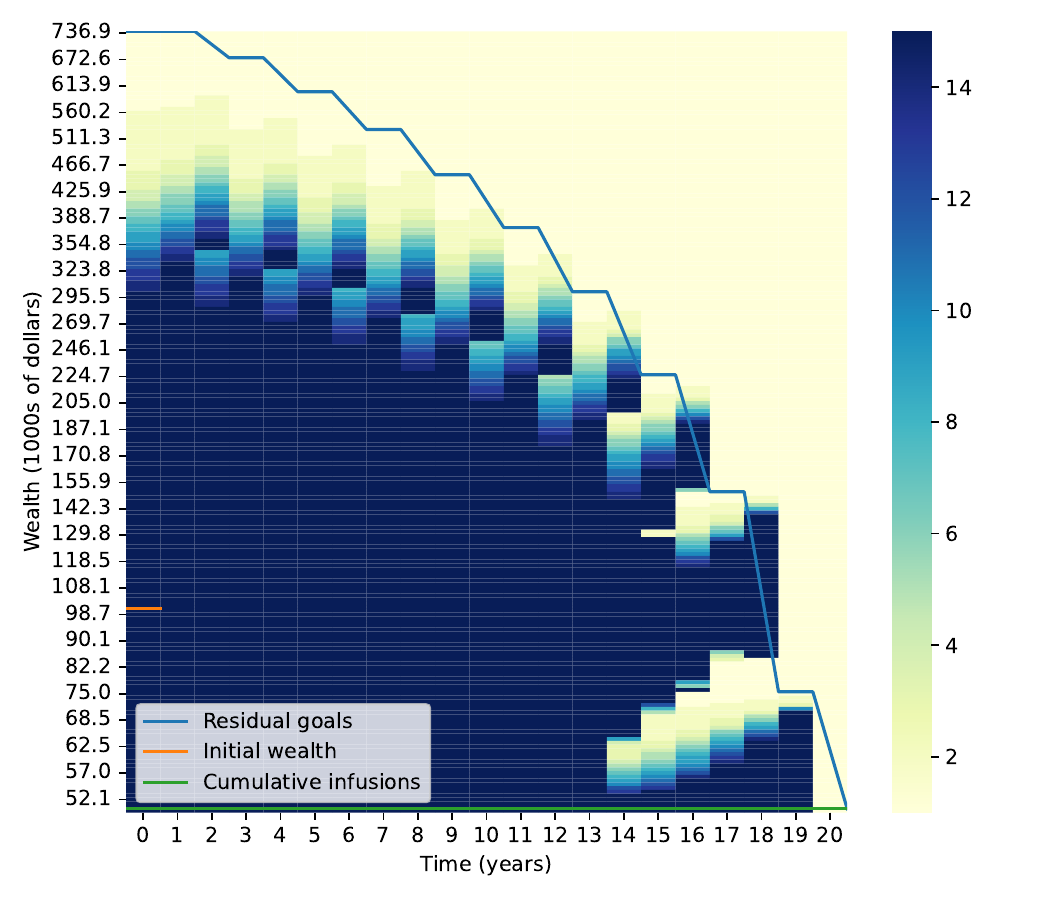}
\includegraphics[scale=0.38]{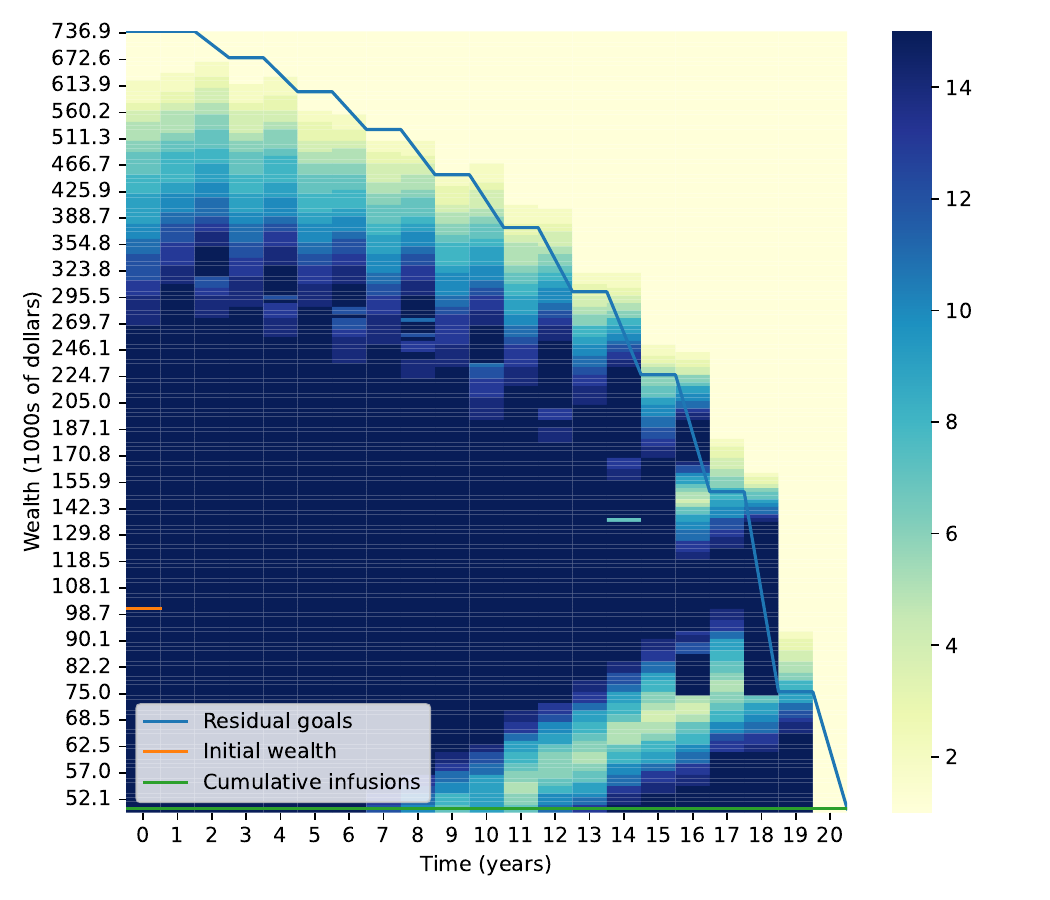}

DP: Goal-taking Decisions \hspace{1.05in}RL Inference: Goal-taking Decisions\\
\includegraphics[scale=0.38]{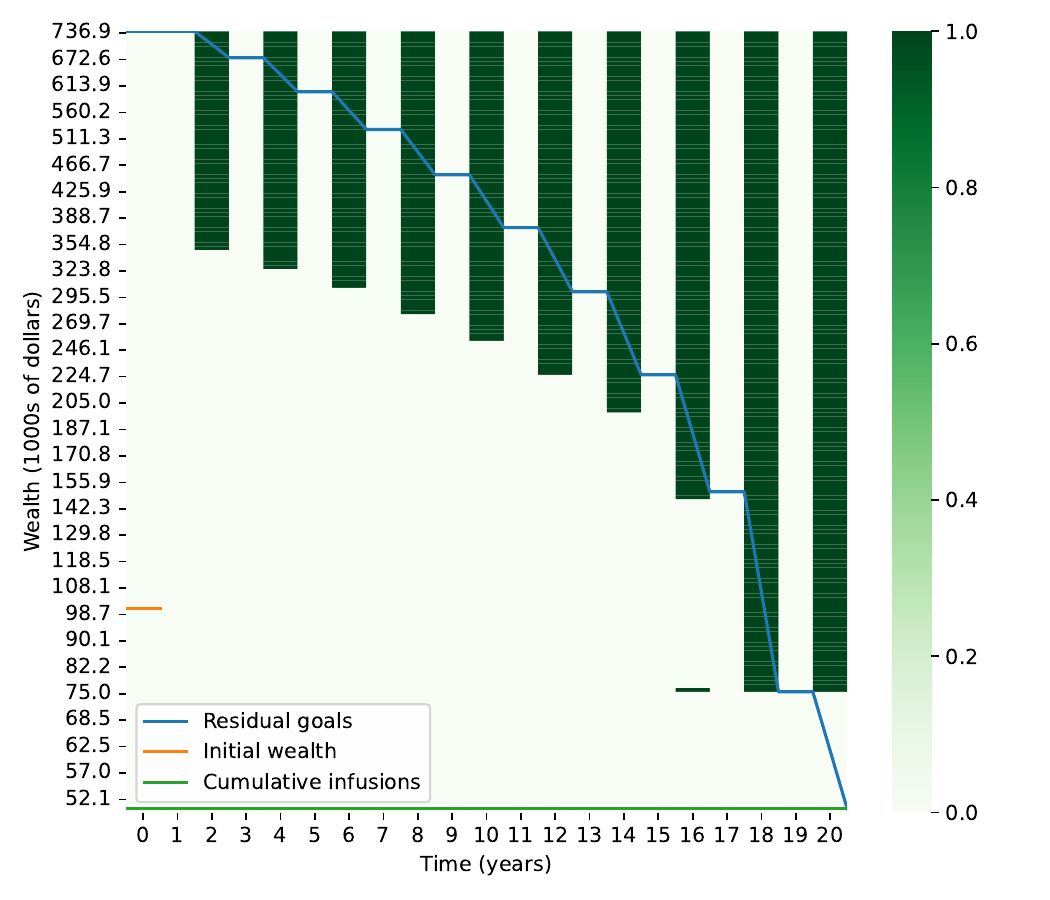}
\includegraphics[scale=0.38]{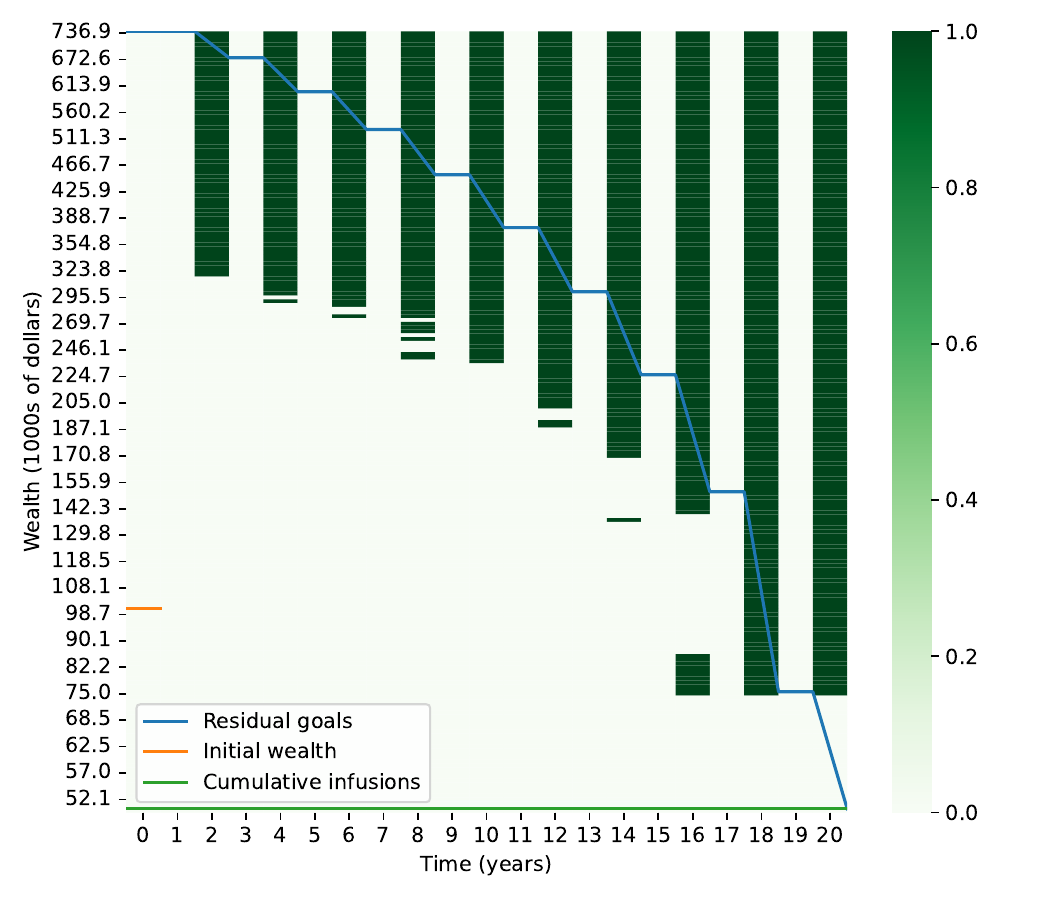}
\caption{\label{fig:policy_heat_maps} \small 
Heatmaps comparing the investment portfolio decisions and goal-taking decisions made by DP versus RL inference for test suite case 20. In the investment portfolio plots, the darker the color, the more aggressive the investment portfolio. In the goal-taking plots, the dark colored bars are regions in which the available goal is taken. The initial wealth, 100 (thousand) dollars, is denoted by the orange bar at the initial time. The jagged blue line denotes the cost to fulfill all the remaining goals.}
\end{figure}

We next look at how the decisions determined using RL inference compare to DP's optimal decisions. DP determines its optimal decisions over a grid in the $(t,W)$ state space plane. We then use RL inference to determine both the goal-taking decision and the investment portfolio decision at each point in the same $(t,W)$ grid. To speed computational time, the MetaRL algorithm is queried in parallel using all the grid's wealth values at once for each given time $t$. For any specific problem, the resulting decisions can be compared by computing heatmaps over the $(t,W)$ grid using both DP and RL inference for comparing either the goal-taking decisions or the investment portfolio decisions. We demonstrate this with a representative example from our 66 test case problems, specifically case 20. From Table \ref{tab:testcases_noinf}  in Appendix \ref{66}, we see that in this case: (1) There are 20 time steps (i.e., 20 years). (2) The initial wealth is 100 (thousand) dollars. (3) At each even time step (i.e., 2,4,...,20), there is a goal available that costs 75 (thousand) dollars, which, if taken, creates a utility of 1 for the investor. (4) There are no infusions. A comparison of the decisions made by RL inference versus DP can be found in Figure \ref{fig:policy_heat_maps}.  In this case the optimal expected attained utility (i.e., the DP value function) is 4.10 according to the DP backwards pass calculation. In the next subsection, we will discuss how we estimate the optimal expected attained utility using the RL heatmaps given here, which will produce an average attained utility of 4.02. 

A second representative example from our 66 test problems is given in Appendix \ref{sec:add_comp}, which leads to similar results. In both examples, we see that the RL inference graphs retain remarkable amounts of details that are shown in the optimal DP graphs. Because of this, it is no surprise that the estimate of the expected attained utility using RL inference is so close to its optimal value as determined by DP. We next show how this estimate is formed.

\subsection{Expected Attained Utility Analysis: Comparing RL Inference To DP}\label{sec:EUA}

%How do we use RL inference to estimate the optimal expected attained utility? And how good is this estimate?

After the process in the previous subsection is run to use the MetaRL algorithm to create decision heatmaps, we run 10,000 Monte Carlo trajectories, corresponding to 10,000 different histories for the value of $Z$ at each time step in the geometric Brownian motion model given in equation \eqref{wealth_transition}. At each time step for each trajectory, the goal-taking decision (if there is one) and the investment portfolio decision are taken from the nearest wealth grid point to the trajectory's wealth at that time in the heatmap. These 10,000 trajectories are run in parallel in each time step to save computational time. The RL inference estimate for the optimal expected attained utility is the average of these 10,000 attained utilities. The RL-Efficiency is this number divided by the attained utility using the optimal decisions determined from DP averaged over the same 10,000 Monte Carlo simulations. In Table \ref{tab:expected_utility_comparison}, we present statistics for the RL-Efficiencies of the 66 test suite problems.

\begin{table}[h!]
\centering
\caption{\label{tab:expected_utility_comparison} 
RL-Efficiencies for the 66 test suite cases in Subsection \ref{sec:test_suite}. The RL-Efficiency essentially gives the attained utility from RL inference over the maximum possible attained utility determined from DP. The closer the RL-Efficiency is to 1, the more optimal RL inference is.}

\begin{tabular}{p{1.59in}|c|c|ccccc}
\toprule
 & mean & std.~dev. & min & 25\% & 50\% & 75\% & max \\
\midrule
RL-Efficiency (from 10,000 Monte Carlo simulations) & 0.978 & 0.016 & 0.917 & 0.975 & 0.983 & 0.987 & 0.999 \\
\bottomrule
\end{tabular}

\end{table}

We note in Table \ref{tab:expected_utility_comparison} that the mean RL-Efficiency of the 66 test suite examples is 0.978. But how stable is this result? In particular, is 10,000 a sufficiently high number of Monte Carlo trajectories? To test this, we computed the mean RL-Efficiency of the 66 test suite examples 30 times using 10,000 simulations each time. The mean of these 30 times was 0.978293 with a standard deviation of 0.000380, meaning 0.978 is a very stable number, and 10,000 Monte Carlo trajectories gives almost three digits of accuracy for the mean RL-Efficiency. This is a particularly strong result given that we are just using RL inference without training on the specific test problems being considered.

%The fact that our mean RL-Efficiency is 0.978 answers the second question given at the beginning of this subsection: The estimate of the optimal expected attained utility starting at the initial time using our RL inference method is quite close to the optimal answer produced by DP. This is a particularly strong result given that we are just using RL inference without training on the specific scenario being considered. There are other advantages of our meta-model approach. The pre-trained MetaRL model can be pushed to thin clients for inference. It also enables distribution of the trained policy functions (parameters and weights) without revealing the code for the training procedure and the data and problems on which the MetaRL model has been trained. This open model distribution mirrors the release of open models in the realm of LLMs. 

\subsection{Investment Analysis: Robustness Of MetaRL To 
Varying The Efficient Frontier}\label{EFront}

\begin{figure}[h!]
    \centering
    \includegraphics[width=0.95\linewidth]{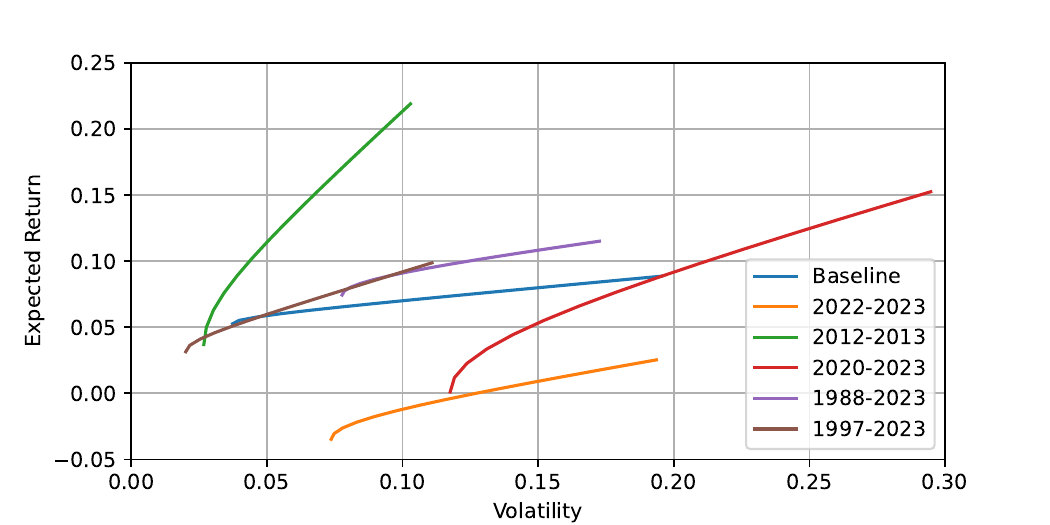}
    \caption{\small The baseline efficient frontier used for training throughout this paper and, other than in Subsection \ref{EFront}, all of the examples in this paper. In Subsection \ref{EFront}, the other five efficient frontiers, labeled by the time ranges of the returns used to generate them, are used for out-of-distribution testing examples.}
    \label{fig:efs}
\end{figure}

All the results previously presented in this section, along with the MetaRL training, have used investment portfolios along the blue ``baseline'' efficient frontier shown in Figure \ref{fig:efs}. This efficient frontier was generated by applying Markowitz mean-variance optimization \citep{markowitz_portfolio_1952} to the returns from a U.S.~stock index, a U.S.~bond index, and an international stock index\footnote{More specifically, (i) Vanguard Total Stock Market Index Fund Investor Shares (VTSMX) was used for the U.S.~stock index, (ii) Vanguard Total Bond Market II Index Fund Investor Shares (VTBIX) was used for the U.S.~bond index, and (iii) Vanguard Total International Stock Index Fund Investor Shares (VGTSX) was used for the international stock index.} between January 1998 and December 2017. The most conservative portfolio investment corresponds to the lowest possible volatility on the frontier. The most aggressive corresponds to the largest expected return of the three indexes, in this case the U.S.~stock index. The $P$ investment portfolios are equally spread in terms of their expected return values between the most conservative and most aggressive cases.

Over time, the governing efficient frontier will change, sometimes considerably. It is therefore important to see if the MetaRL model can accommodate the wide range of efficient frontiers and investment portfolios that it might encounter over time. To test this, we consider the effect on the mean RL-Efficiency when the efficient frontier is changed for all 66 test suite examples, even though the MetaRL model is trained strictly on the baseline efficient frontier discussed above. 

For this robustness test, we considered five different efficient frontiers labeled in Figure \ref{fig:efs} by the varying ranges of years of returns used to generate them. They also varied in terms of combinations of the underlying investments, using returns from different combinations of the three indexes above, along with a large-cap value U.S.~stock fund, a large-cap growth U.S.~stock fund, a value international stock fund, and a long-term U.S.~bond fund. In all cases, the choices for the five efficient frontiers were guided by looking for unusual efficient frontiers with interesting behavior in both good and bad markets to push our stress test as far as realistically possible.

For our the baseline case, the ratio of the optimal expected attained utility over the total utility if all the goals are attained, averaged over the 66 test suite examples, is 0.636. If we use a different efficient frontier with, say, a higher expected return that causes this average ratio to be close to 1, then we expect to attain most of the utility even if we use a somewhat suboptimal investment portfolio strategy, meaning the mean RL-Efficiency is artificially pushed closer to 1. To prevent this artificial boost, for each efficient frontier, we scaled the initial wealth of the 66 examples so that this average ratio stayed between 0.63 and 0.64. 

With the initial wealths correctly scaled, we computed the mean RL-Efficiency for each of the efficient frontiers. These are given in Table \ref{EFRes}. Each of the new efficient frontiers actually shows a slight improvement from the baseline case mean RL-Efficiency value of 0.978. This appears to be due to the baseline efficient frontier showing less of an increase in the expected return as the volatility increases, but the results certainly indicate that the MetaRL model is quite robust to changes in the efficient frontier. Thus, the MetaRL solution developed in one economic regime is remarkably robust to regime shifts. This is a key result since regime shifts are both inevitable and unpredictable.

\begin{table}[h!]
\centering
\caption{\label{EFRes} \small
RL-Efficiencies for the 66 test suite examples in Subsection \ref{sec:test_suite}, using the wide range of the six efficient frontiers shown in Figure \ref{fig:efs}. Even though the MetaRL model is trained on the baseline case, the RL-Efficiencies for the other five efficient frontiers are better.}
{\footnotesize
\begin{tabular}{p{2in}|c|c|}
\toprule
Efficient Frontier & Mean RL-Efficiency \\
\midrule
 Baseline &   0.978 \\
 2022--2023 & 0.986 \\
 2012--2013 & 0.995 \\
 2020--2023 & 0.990 \\
 1988--2023 & 0.984 \\
 1997--2023 & 0.987 \\
\bottomrule
\end{tabular}
}
\end{table}

A key technical reason for the observed robustness is that the core MetaRL policy operates in a fundamentally stationary and normalized state space. For each experiment, the efficient frontier is first computed using the most recent available information and then discretized into a finite set of portfolios. The RL agent does not act directly on raw asset returns; instead, its state inputs are fully normalized and frontier-agnostic. The policy therefore learns to map normalized goal and wealth states to a position along a generic efficient frontier, rather than to a specific set of assets or a particular return distribution. At inference time, the chosen normalized portfolio is simply de-normalized to select the corresponding point on the prevailing frontier.

This design makes the trained policy largely oblivious to the underlying return regime. Changes in macroeconomic conditions or asset combinations affect the shape of the frontier, but not the structure of the decision problem presented to the agent. Moreover, as the separation between portfolios becomes more statistically significant (i.e., steeper or more dispersed frontiers), the discrimination problem becomes easier, which improves measured efficiency. Hence, the MetaRL solution transfers naturally across regimes because it learns a generalized allocation rule on the frontier itself, rather than overfitting to any particular economic environment. 

\section{Extensions Of The MetaRL Model}
\label{sec:extensions}

\subsection{Incorporating Concurrent And Partial Goals}
\label{subsec:concurrent_partial_goals}

Up to this point, we have restricted ourselves to at most one all-or-nothing goal each year. In Appendix \ref{PCG}, we show how the MetaRL model can be extended to more than one concurrent goal and to allowing some or all goals to be partially filled. In this case of ``partial goals'' the investor may ideally want the full goal, like a luxury trip to Paris for a month, but may also be open to partial goals like only having two weeks in Paris or having a month in New Jersey instead. Each partial goal has a reduced cost from the full goal, but also has a reduced utility if it is chosen. Dynamic Programming can also be used with concurrent or partial goals (see \citet[]{das_dynamic_2022} or \cite{capponi_continuous_2023}, for example), so we can continue to compare the RL inference results to DP, as we did in Subsections \ref{sec:DA} and \ref{sec:EUA}. 

We consider a test suite of 4 examples here, based off the cases given in \citet[]{das_dynamic_2022}. The first case (CP1) is identical to the case used in Subsection 4.3.1 of 
\citet[]{das_dynamic_2022}. In this case there are three concurrent goals at time 5: one all-or-nothing goal (so there are two possibilities since the investor can either take or not take the goal), one goal with one partial goal (so three possibilities: take the full goal, take the partial goal, or do not take the goal), and one goal with three partial goals (so five possibilities), and at time 10 there is one goal that has one partial goal. We note that this gives $2\times3\times5\times3=90$ different goal-taking combinations to choose from over time. The second case (CP2) is identical to the case used in Subsection 4.5 of \citet[]{das_dynamic_2022}. This is a far more complicated scenario involving over 301 full goals and 138 partial goals to choose from over the course of $T=60$ years. The third case (CP3) only uses the car goals from the previous example. These occur every five years, and each allows for a fancy car (full goal), a less fancy car (partial goal), or no car to be considered. The final case (CP4) is a more complicated variation of CP3. It also allows for a fancy car (at a cost of 32 (thousand) dollars for a utility of 125) or a less fancy car (at a cost of 22 (thousand) dollars for a utility of 110) to be purchased every five years (i.e., at $t= 5, 10,...,55$ years) but it also considers purchasing a trip every ten years (i.e., at $t= 10, 20,...,50$ years) at a cost of $\left(\frac{t}{2} + 10\right)$ thousand dollars for a utility of 100. For CP4, we start with 25 (thousand) dollars and add (infuse) 1 (thousand) dollars every year.

Table \ref{tab:CP} shows the results for each of these four cases. The first three columns of the table look at the computational time. While these examples with concurrent and partial goals generally take more time than the all-or-nothing goals, which is no surprise, the more interesting information is in the time ratio, where we see that as the test suite example gets more complicated, RL inference takes progressively less time to produce a solution compared to DP. Further, from the fourth column, which is for RL-Efficiency, we see that the additional complexity of the example does not necessarily mean a degradation in the quality of the solution, noting that the worst RL-Efficiency, which is for case CP4, is the second least complex case.

\begin{table}[h!]
\centering
\caption{\label{tab:CP} \small
Speed (measured in seconds) and accuracy (measured by the RL-Efficiency) for RL inference versus DP for each of the four cases with concurrent and/or partial goals in our new test suite}
{\footnotesize
\begin{tabular}{c|cccc}
\toprule
Test suite case   & DP time (in sec) & RL inference time (in sec) & DP/RL time ratio & RL-Efficiency \\ \midrule
CP1 & 0.276         & 0.271         & 1.02   & 0.990       \\
CP2 & 14.6         & 4.50         & 3.24   & 0.985      \\
CP3 & 2.94         & 1.48         & 1.98   & 0.986      \\
CP4 & 3.36         & 1.54         & 2.18   & 0.972     \\
\bottomrule
\end{tabular}
}
\end{table}

\begin{figure}[ht!]
\centering
DP: Investment Portfolio Decisions \hspace{.55in}RL Inference: Investment Portfolio Decisions\\
\includegraphics[scale=0.38]{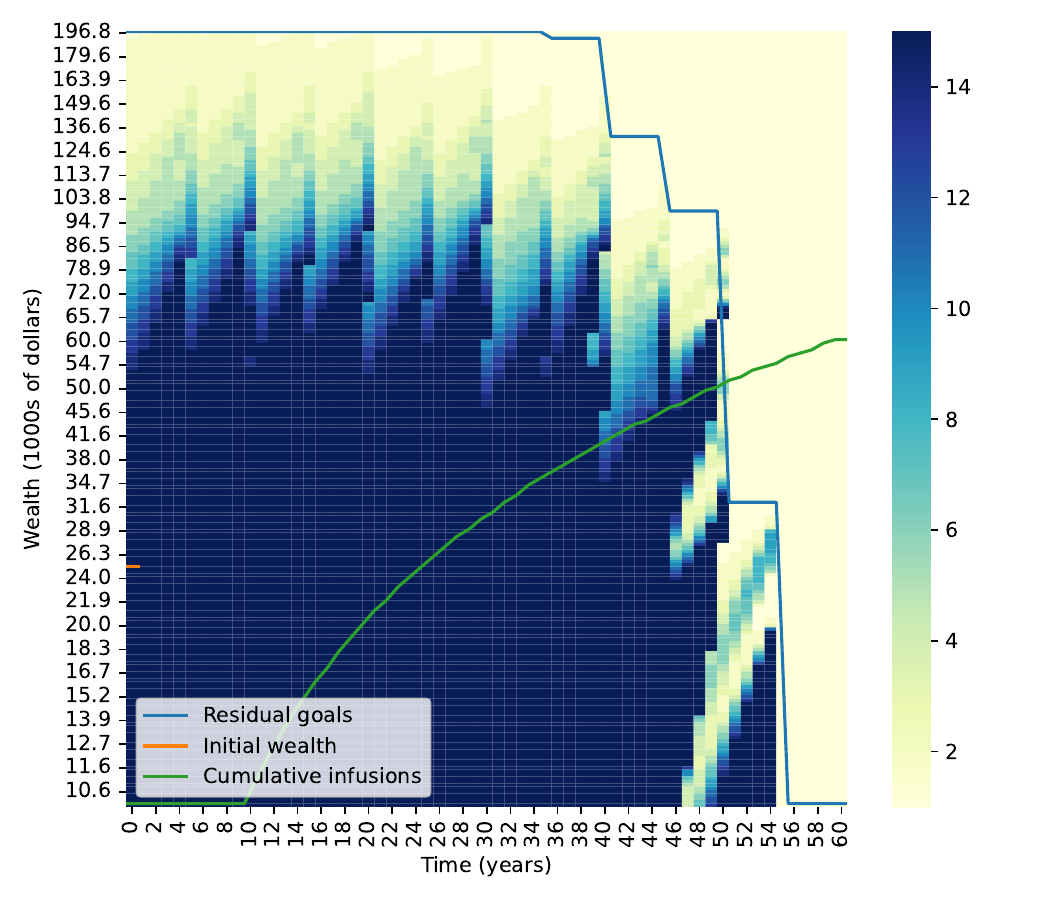}
\includegraphics[scale=0.38]{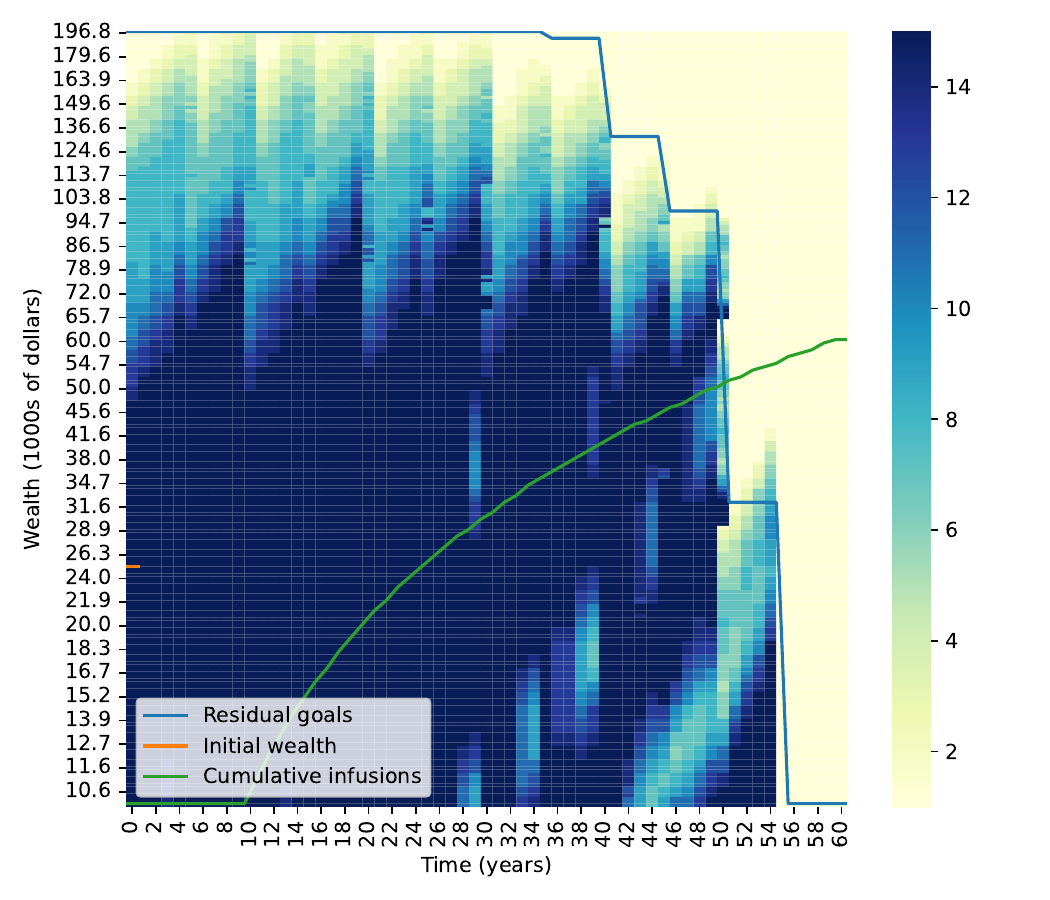}

DP: Goal-taking Decisions \hspace{1.05in}RL Inference: Goal-taking Decisions\\
\includegraphics[scale=0.38]{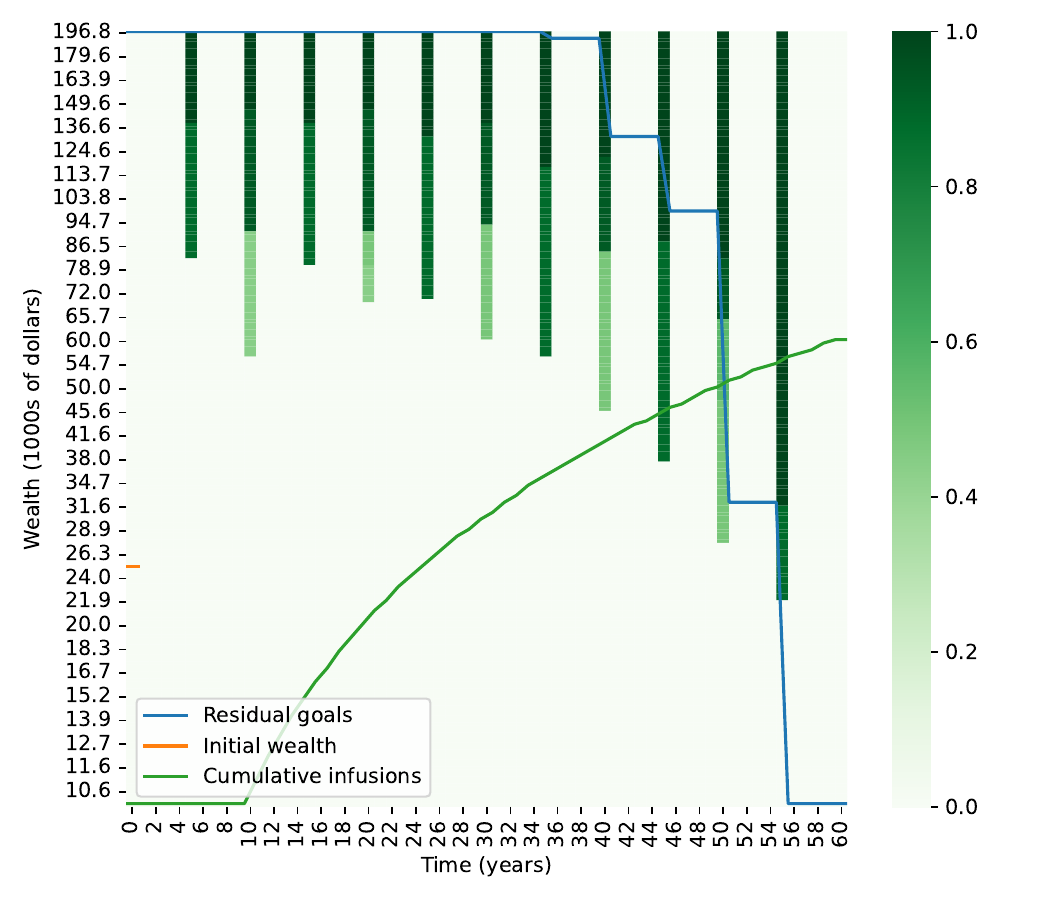}
\includegraphics[scale=0.38]{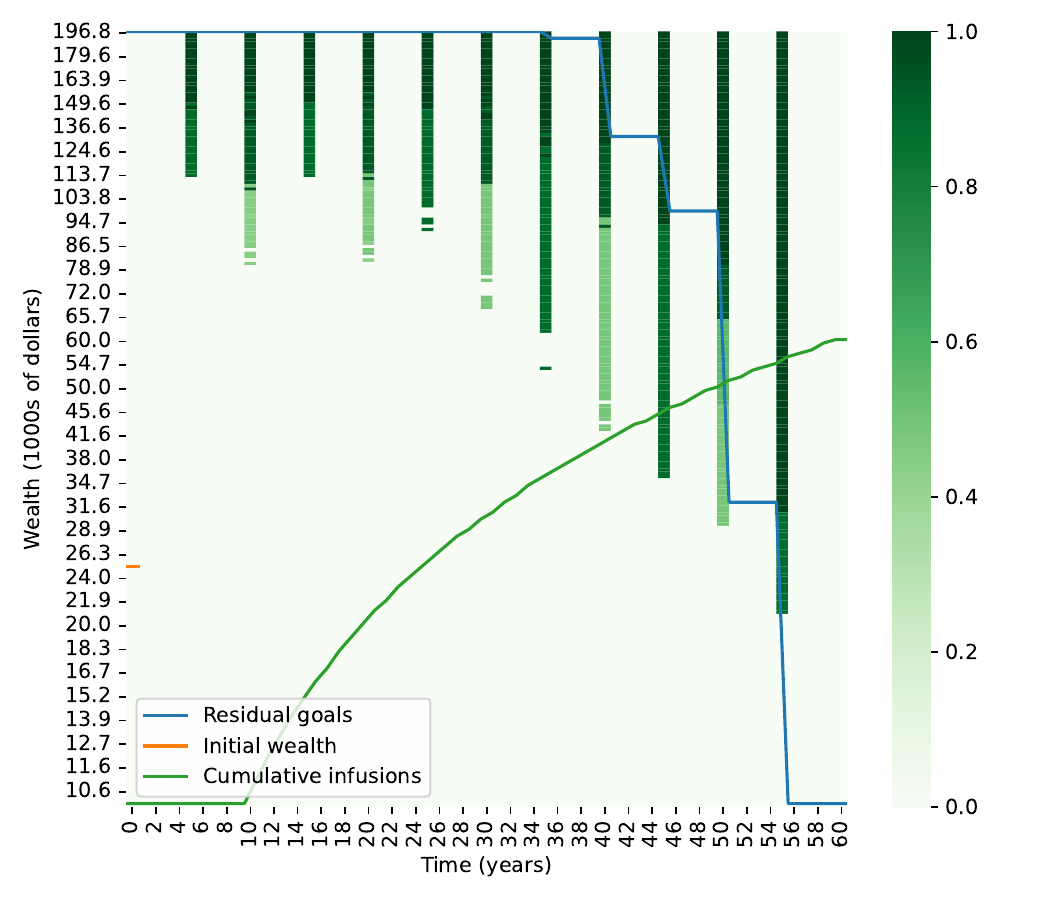}

DP: Goal proportions \hspace{1.25in}RL Inference: Goal proportions\\
\includegraphics[scale=0.38]{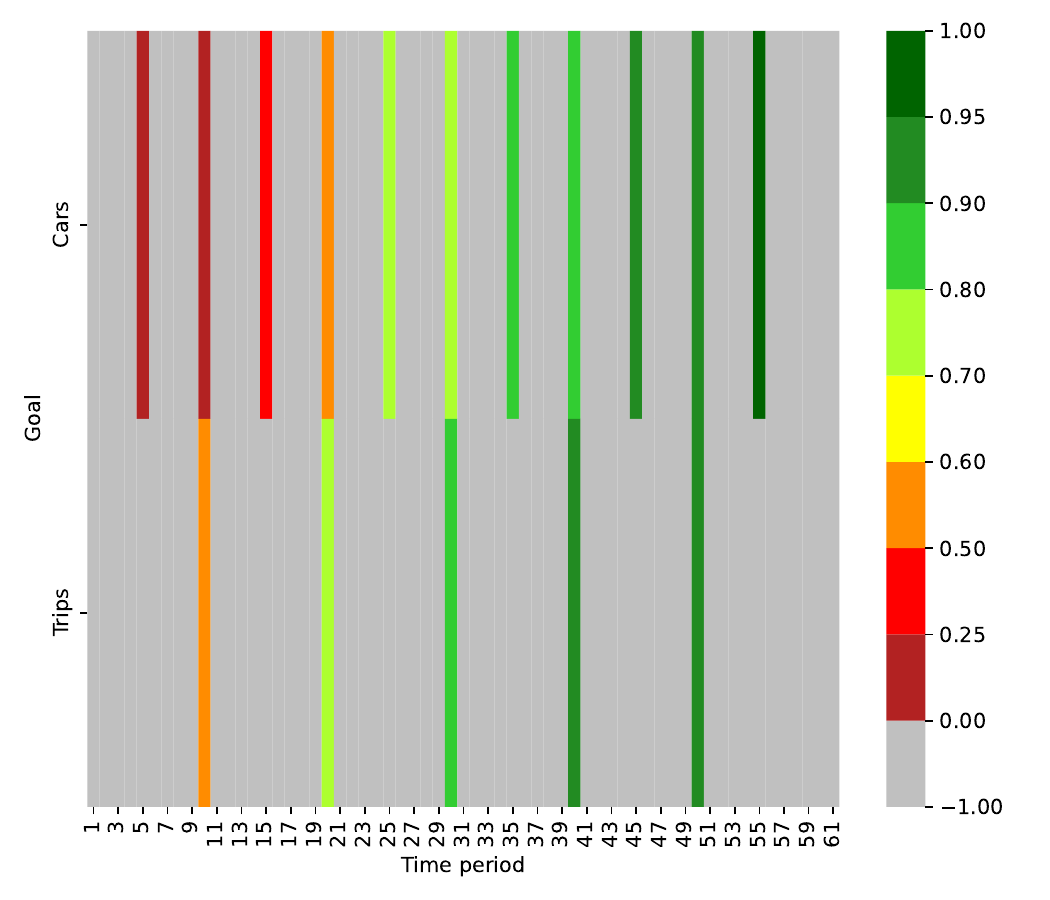}
\includegraphics[scale=0.38]{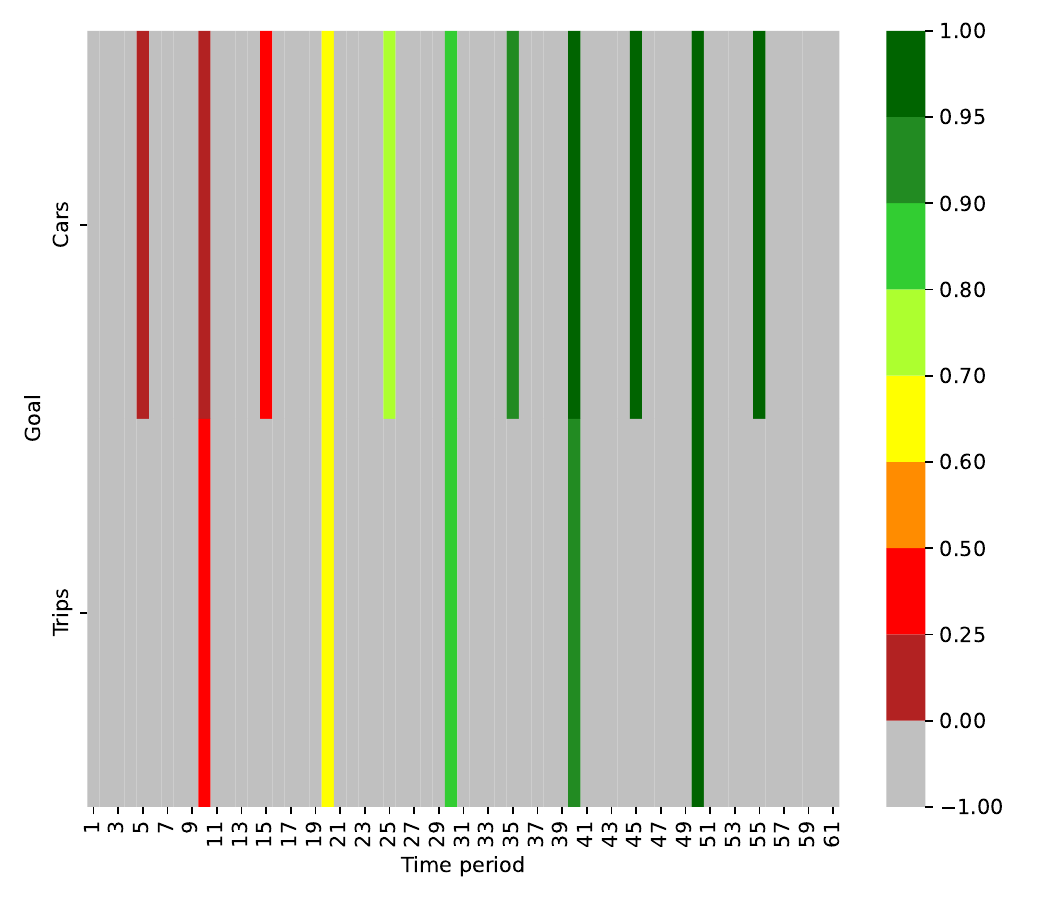}
\caption{\label{fig:policy_heat_maps_part_conc} \small Heatmaps for investment portfolio decisions and goal-taking decisions with case CP4, which has both concurrent goals (cars and trips) and partial goals (for both cars and trips). The specifics of case CP4 are in the text. The colors in the second row (Goal-taking Decisions) indicate the ratio of the expected attained utility from that year's goals over the utility attained if all of that year's full goals were fulfilled. The graphs in the bottom row correspond to this same ratio, but for each individual goal (i.e., for cars and for trips).
}
\label{CPfun}
\end{figure}

Given that CP4 has the lowest RL-Efficiency, we choose to look at the heatmaps for CP4's RL inference decisions in relation to DP's, since a lower RL-Efficiency indicates the potential for a larger difference between the DP and RL inference heatmaps. Figure \ref{CPfun} compares these heatmap results. Figure \ref{CPfun}'s format largely parallels the format in Figures \ref{fig:policy_heat_maps} and \ref{fig:policy_heat_maps_2}, however since the goal-taking each year is no longer all-or-nothing, we introduce colors in the second row of graphs to indicate the fraction of the average utility that is attained over the utility attained if all the full goals were fulfilled that year. We also have added the bottom row to indicate this same fraction for each goal individually. For example, in the DP graph in the bottom row, the colors indicate that in year 20, the expected utility from the car goal is 50-60\% of the maximum possible utility attained if the full car goal were fulfilled, and this number jumps up to 70-80\% for the trip goal. Figure \ref{CPfun} largely shows that the goal-taking decisions between DP and RL inference are quite similar, while the investment portfolio decisions have the same qualitative behavior but differ in that RL inference tends a bit towards more aggressive investment portfolios than DP. 

\subsection{Incorporating Stochastic Inflation}
\label{sec:stochastic_inflation}

Due to what is the commonly referred to as the ``curse of dimensionality,'' Dynamic Programming, from a computational point of view, is generally limited to 2 or 3 state variables. For any GBWM model that also addresses the effect of stochastic inflation, we have our two previous state variables, the current wealth and the current time, along with two new state variables, the current inflation and, to update real infusions and the costs of real goals, the cumulative effect of inflation up to the current time must also be taken into account. That is, we have 4 state variables, meaning that Dynamic Programming is not a computationally feasible method if we wish to incorporate stochastic inflation.

MetaRL, on the other hand, easily extends to incorporate stochastic inflation models. As an example, we consider the \cite{vasicek_equilibrium_1977} model for inflation, which is governed by the stochastic differential equation
\begin{equation}\label{SDE1}
dI_{\mathrm{infl}} = -\kappa_{\mathrm{infl}}(I_{\mathrm{infl}} - \theta_{\mathrm{infl}})dt+\sigma_{\mathrm{infl}} dW,
\end{equation}

\noindent where $W$ is a Wiener process. This model is a mean reverting process, where $\theta_{\mathrm{infl}}$ is the mean value of $I_{\mathrm{infl}}$, $\kappa_{\mathrm{infl}}$ is the constant strength of the reversion to this mean value, and $\sigma_{\mathrm{infl}}$ is a volatility constant representing the strength of the randomness. The method for incorporating the current inflation and the cumulative effect of the inflation is contained in Appendix \ref{sec:app_stoch_infln}, which we note uses 27 state variables. Experiments show the changes given in Appendix \ref{sec:app_stoch_infln} have {\it no} discernible impact on runtimes, either for training the MetaRL model or for running inferences with it. 

\begin{table}
\centering
\caption{\small The effect of inflation as we vary the three inflation parameters, $\theta_{\mathrm{infl}}, \kappa_{\mathrm{infl}},$ and $\sigma_{\mathrm{infl}}$, in the Vasicek model. The table shows the fraction of expected attained utility in the presences of stochastic inflation over the expected attained utility when there is no inflation. In all cases, the initial inflation rate is set equal to the mean, $\theta_{\mathrm{infl}}$.  
}\vspace*{0.1in}
\label{tab:kappa_theta_sigma}
\begin{tabular}{c|ccccc}
\multicolumn{6}{c}{Panel A: $\kappa_{\mathrm{infl}}=0.1$} \\ \midrule
      & \multicolumn{5}{c}{$\sigma_{\mathrm{infl}}$} \\ 
$\theta_{\mathrm{infl}}$ & 0.00      & 0.01     & 0.02     & 0.03     & 0.06     \\ \hline

0.01 & 0.943 & 0.940 & 0.931 & 0.907 & 0.804 \\
0.02 & 0.883 & 0.878 & 0.876 & 0.869 & 0.783 \\
0.03 & 0.821 & 0.817 & 0.818 & 0.823 & 0.764 \\
0.05 & 0.701 & 0.706 & 0.713 & 0.727 & 0.724 \\
0.08 & 0.601 & 0.609 & 0.615 & 0.623 & 0.666 \\
0.10 & 0.601 & 0.599 & 0.595 & 0.598 & 0.632
\\
\bottomrule
\end{tabular}
\mbox{}\vspace{0.25in}
\begin{tabular}{c|ccccc}
\multicolumn{6}{c}{Panel B: $\kappa_{\mathrm{infl}}=0.6$} \\ \midrule
      & \multicolumn{5}{c}{$\sigma_{\mathrm{infl}}$} \\ 
$\theta_{\mathrm{infl}}$ & 0.00      & 0.01     & 0.02     & 0.03     & 0.06     \\ \hline
0.01 & 0.943 & 0.943 & 0.942 & 0.942 & 0.938 \\
0.02 & 0.883 & 0.883 & 0.882 & 0.881 & 0.877 \\
0.03 & 0.821 & 0.821 & 0.820 & 0.819 & 0.817 \\
0.05 & 0.701 & 0.702 & 0.703 & 0.704 & 0.708 \\
0.08 & 0.601 & 0.600 & 0.601 & 0.606 & 0.610 \\
0.10 & 0.601 & 0.600 & 0.600 & 0.600 & 0.595
\\
\bottomrule
\end{tabular}
\end{table}

Table \ref{tab:kappa_theta_sigma} shows the effect that stochastic inflation has on investors' utility when averaged across the 66 test problems used in Section \ref{sec:test_suite}. More specifically, the table contains the average optimized utility over all 66 test cases (each using 10,000 Monte Carlo paths) in the presence of stochastic inflation and divides this by 481.54, which is the average optimized utility with the same cases and paths if there is no inflation. In all cases, the initial inflation is set equal to the mean inflation, $\theta_{\mathrm{infl}}$. 

One effect that is clear from the tables is that the higher the inflation, the lower the expected attained utility. This is simply a ramification of the fact that the higher the inflation, the higher the costs of the goals become (since they increase with inflation), so fewer goals are financially attainable. 

The more important question is what is the effect of the inflation being stochastic, as opposed to constant? If the inflation can be assumed to be constant (i.e., equal to $\theta_{\mathrm{infl}}$), then the goals' costs can be adjusted for inflation at time $=0$, and we can revert back to our previous non-stochastic inflation analysis if desired. Our MetaRL approach displayed in Table \ref{tab:kappa_theta_sigma} shows that when $\kappa_{\mathrm{infl}}$, the strength of the mean reversion, is a weak value like 0.1 in Panel A, the effect of the inflation being stochastic (seen by looking at the variation over a row in the table) can be significant. However, when the strength of the mean reversion is a stronger value like 0.6 in Panel B, there is generally little variation across the rows and it is reasonable to treat the inflation as constant.

Looking at historical inflation in the United States for example, we can approximate $\theta_{\mathrm{infl}} = 0.03, \kappa_{\mathrm{infl}}=0.6$, and $\sigma_{\mathrm{infl}}=0.03$. From Panel B where $\kappa_{\mathrm{infl}}=0.6$, we see that when $\theta_{\mathrm{infl}} = 0.03$, the difference between using the average inflation (i.e., $\sigma_{\mathrm{infl}}=0$) versus $\sigma_{\mathrm{infl}}=0.03$ is trivial. In fact, choosing $\theta_{\mathrm{infl}} = 0.031$ makes about three times more of a difference to the expected accumulated utility than including the effect of inflation being stochastic. Therefore, our MetaRL analysis shows that using dynamic programming with the average rate of inflation instead of stochastic inflation is justified for current calculations in America---a conclusion that could not be found through dynamic programming. In each new $(\theta_{\mathrm{infl}}, \kappa_{\mathrm{infl}}, \sigma_{\mathrm{infl}})$ scenario for other countries or different circumstances, our MetaRL analysis can be used to determine if using the average inflation is justified or, if not, how to accommodate stochastic inflation.

\section{Discussion of Related Literature}\label{sec:litdetailed}

The work in this paper comes after decades of research on how people should save, consume, and invest. Below we provide a detailed review of the antecedent and related literature, which may be read in the context of the preceding sections. While brief, it depicts how this paper stands on the shoulders of many good ideas in past work. We review continuous-time dynamic consumption investment models and regime-based models, solved with traditional methods and more recently, with RL. We also discuss the evolution of the goals-based wealth management paradigm, the main subject of this paper. These problems are related to asset-liability management and stochastic programming. We also mention the MetaRL literature and techniques like PPO, and newer work related to finance. 

Dynamic optimization of investment portfolio decisions over time was introduced in \cite{merton_lifetime_1969} and \cite{merton_optimum_1971}. These papers helped establish the theoretical underpinnings for life-cycle investing and consumption-portfolio decisions. They also laid the foundations for optimal lifetime investment portfolio selection under uncertainty using a continuous-time framework. Merton showed how consumption and investment portfolio choice should adapt over the life-cycle in response to changes in the state space (wealth and time). Wealth management decisions also change with employment status, health status, etc.

Several additional papers comprise this literature (e.g., \cite{browne_risk_1999, guidolin_asset_2007, infanger_chapter_2008, topaloglou_dynamic_2008, vo_dynamic_2013, singhal_dynamic_2019}). Dynamic portfolio strategies with regime shifts have also been modeled (e.g., \cite{ang_international_2002, zhou_markowitzs_2003, ang_how_2004, brandt_simulation_2005, brandt_dynamic_2006, guidolin_asset_2007, bernhart_asset_2011, bulla_markov-switching_2011, grobys_active_2012, vo_dynamic_2013, jiang_international_2015, dapena_risk_2019, singhal_dynamic_2019, bellalah_long_2020, lewin_portfolio_2020}; \citet[]{das_optimal_2022}). Examples of other AI applications in finance are in papers that research insurance markets \citep{zhang_ai-powered_2023}, explainable financial anomalies \citep{sabharwal_extending_2024}, and credit scoring \citep[]{chang_prediction_2024, das_credit_2023}, to name a few. 

% Similar techniques have been applied to pension planning problems as well, and more generally to liability-based plans. \cite{dempster_global_2003} explored using dynamic stochastic optimization techniques to find optimal portfolio and contribution strategies for defined contribution pension plans over long time horizons under changing market and economic conditions. This allowed for more customized and risk-controlled approaches. An extensive literature deals with asset-liability management in papers such as  \cite{mulvey_stochastic_1992,  consigli_dynamic_1998, hu_essays_2019}.

The target date fund or life-cycle fund concept aims to mimic optimal portfolio allocations over the life-cycle by adjusting risk exposure (stocks vs.~bonds) as retirement approaches. That is, its investment portfolio strategy is based solely on the time until an investor retires, and nothing else about the investor. Subsequent research has found significant shortcomings in such a simple approach. For example, \cite{duarte_simple_2021} and \cite{duarte_machine_2024} used machine learning algorithms to solve for optimal investment portfolio choices over the life-cycle, accounting for variations in factors like wealth, business cycles, stock valuations etc. They found substantial gains from using customized rules, rather than simple ``age-based'' rules. 
% In \cite{van_binsbergen_optimal_2016}, the impact of ex ante (preventive) and ex post (punitive) risk constraints on dynamic strategies is studied, with the latter being better at resolving underfunding problems in liability plans. 

Two decades ago, \cite{chhabra_beyond_2005} proposed a more holistic wealth allocation framework that integrated factors like human capital, real estate, health, and mortality risk in addition to just financial assets. This provides a more comprehensive view of lifetime risks. This line of work eventually lead to the approach known as goals-based wealth management (GBWM). \cite{brunel_goals-based_2015} provides an overview of goals-based wealth management and proposes changes to traditional wealth advisory practices. 

A series of papers \citep[]{deguest_introducing_2015, dempster_planning_2011, wang_portfolio_2011, pakizer_goals_2017, kim_personalized_2020, martellini_securing_2020, parker_allocation_2020, parker_goals-based_2021,  mohammed_embracing_2021, das_goals-based_2018, das_dynamic_2020, das_efficient_2023}
% das_dynamic_2022, 
% , capponi_continuous_2023
introduced GBWM to the retirement and pension planning industries. Stochastic Dynamic Programming for multi-stage problems been used to obtain solutions to many related asset-liability management optimization problems in papers such as \cite{mulvey_stochastic_1992}, \cite{dempster_calm_1998}, \cite{consigli_dynamic_1998}, \cite{mulvey_financial_2004}, \cite{infanger_chapter_2008}, and \cite{topaloglou_dynamic_2008}.
% \cite{das_dynamic_2022} and \cite{capponi_continuous_2023}
But solving these large-scale, long-horizon optimization problems poses interesting computational difficulties as the state space grows.

\citet[]{das_dynamic_2022} and \cite{capponi_continuous_2023} develop Dynamic Programming methodology for goals-based wealth management that seeks to maximize outcomes over multiple investor goals like retirement, home purchase, education, etc. This approach uses individual investor preferences about their goals to determine the optimal dynamic strategy for when to fulfill and forgo goals and what investment portfolio should be selected as time changes. The approach can be computed quickly, even for many goals over different times, where goals may allow full or partial fulfillment. It computes the probabilities of attaining each goal fully or partially under the optimal dynamic strategy, so investors can ensure their preferences are accurately reflected. It significantly outperforms both target-date funds and commonly used Monte Carlo-based static goal-taking and investment portfolio strategies.

Reinforcement learning (RL) has become popular recently. The classic book by \cite{sutton_reinforcement_2018} introduces RL concepts like Markov decision processes and common RL algorithms like value-based and policy-based methods. The survey paper \cite{hambly_recent_2021} reviews recent developments and applications of reinforcement learning in finance, including for optimal execution \citep{zheng_optimal_2023, nevmyvaka_reinforcement_2006}, portfolio optimization \citep{jiang_deep_2017}, option pricing, market making, order routing, and robo-advising, see also \cite{osterrieder_primer_2023}.

RL is also being used for dynamic optimization of goals-based portfolios. \cite{das_dynamic_2020-1} leveraged seminal developments in papers on Reinforcement Learning, such as \cite{sutton_learning_1988}, which introduced the temporal-difference learning algorithm, a breakthrough in prediction and control methods for reinforcement learning, and \cite{watkins_learning_1989}, which introduced Q-learning, one of the most widely used model-free reinforcement learning algorithms.  \cite{dixon_g-learner_2020} and \cite{halperin_distributional_2021} propose a reinforcement learning approach called G-Learner for goals-based wealth management problems like retirement planning or target-date funds, and GIRL, an inverse reinforcement learning algorithm that can infer the reward parameters of a G-Learner agent from observed behavior data, allowing it to mimic the G-Learner. 

Various algorithms have been applied in RL, some based on deep learning (also known as DeepRL).  \cite{mnih_playing_2013, mnih_human-level_2015} solved optimal playing of Atari games, demonstrating the successful application of deep neural networks as function approximators in reinforcement learning, leading to the rise of deep reinforcement learning. These works showed that a deep reinforcement learning agent could achieve human-level performance across many Atari games, culminating in the beating of the Go  champion, Lee Sedol \citep{silver_mastering_2016}.

Dynamic Programming (DP) (see \cite{bellman_theory_1952} or \cite{bellman1966dynamic}), where each problem has to be re-solved individually, is exemplified in papers such as \cite{brandt_simulation_2005} and \cite{das_dynamic_2020}. Solving using inference with a meta-model is much faster than solving from scratch. This is especially the case as the state space grows, which slows down DP so much that it quickly becomes computationally impossible, while barely slowing the MetaRL approach at all.

The MetaRL model in this paper comprises deep neural net policy functions that take in the values of state variables and return the concomitant policy actions. 
% The algorithms return RL value functions that are used to evaluate the quality of the policy functions. The RL value functions are also deep neural networks that take state variables as inputs. 
We are in a class of RL algorithms known as ``actor-critic'' models, where the policy function is the actor and the RL value function is the critic used to evaluate the quality of the policy function. \cite{sutton_learning_1988} was the first to formally introduce and analyze the actor-critic framework. This was improved in the paper by \cite{schulman_trust_2015} and \cite{schulman_proximal_2017}, which introduced the Proximal Policy Optimization (PPO) algorithm, a widely used policy gradient method for reinforcement learning. \cite{haarnoja_soft_2018} proposed the ``soft actor-critic'' algorithm, a maximum entropy reinforcement learning method that is robust and sample-efficient. We implement a variant of the PPO algorithm in this paper with a separate actor and critic for the goal-taking strategy and for the investment portfolio strategy of an investor. Hence, there will be two policy functions and two RL value functions, all trained simultaneously, i.e., a dual actor-critic formulation.

The MetaRL model mimics ideas in large language models (LLMs), which are trained on large text corpora (e.g., \cite{brown_language_2020, wei_finetuned_2022, izacard_atlas_2022, kojima_large_2023,  gemma_team_gemma_2024}).\footnote{These models are closed (only API access) or they may be open (some of the code, training data, and model weights are publicly available).} LLMs are pre-trained and may be applied to text tasks without re-training the model. Similarly, the MetaRL model in this paper is trained on thousands of GBWM investor problems under a variety of financial circumstances. We implement our dual-PPO algorithm on thousands of investor problems to train a MetaRL model. Analogous to the use of LLMs in inference mode, MetaRL provides zero-shot answers to both investment portfolio and goal-taking decisions. Additionally, the approach in this paper is inspired in part by work such as \citep{finn_model-agnostic_2017, gupta_meta-reinforcement_2018, finn_meta-learning_2018, humplik_meta_2019, lin_model-based_2022}, and for a survey, see \cite{beck_survey_2023}, which proposed the use of a meta-model to handle tasks with unseen environment parameters (such as different financial outlooks and investment portfolios in our case). These papers have made significant contributions to the field of meta reinforcement learning, introducing key algorithms, theoretical foundations, and novel approaches for enabling fast adaptation and generalization with reinforcement learning across different tasks and environments.

\section{Concluding Comments}
\label{sec:conclusion}

This paper shows a new approach that combines advanced computational techniques with a comprehensive financial planning framework to significantly enhance the scope, speed of computing, and effectiveness of wealth management strategies. We develop a meta-model based on reinforcement learning (RL) called MetaRL, which is pre-trained on thousands of goals-based wealth management (GBWM) problems. This model provides rapid, near-optimal solutions for new investor problems without the need for any re-training, delivering expected utilities that closely approximate those obtained through Dynamic Programming (DP) separately solving each problem. More specifically, the ``RL-Efficiency'' (i.e., the ratio of the estimated optimal expected attained utility from using inference with MetaRL to the same result if we use the optimum decisions obtained from DP) is shown to be, on average, 97.8\% across a suite of test problems that have a wide range in initial wealth values, investor horizons (up to 100 years), infusions, goals (from only one goal at the horizon to goals every year) and goal properties (timing, costs, and utilities). 

%Rather than training a new model for each new investor's retirement planning problem, the MetaRL approach invests in a one-time pre-training effort to create a model that can then be called directly for a new investor problem, thereby offering essentially instant inference to determine near-optimal goal-taking strategies and investment portfolio decisions. Thus, MetaRL leverages ideas from the pre-training of foundation models such as LLMs. MetaRL can be deployed on thin clients for inference. Its parameters and weights can be released without revealing the code for the training procedure and the data and problems on which the MetaRL model has been trained, mirroring the release of open models in the realm of LLMs. 

MetaRL efficiently addresses the complexities inherent in multi-goals wealth management. By generating policy decisions within hundredths of seconds that are optimized both for dynamic goal-taking decisions and for dynamic investment portfolio selection, MetaRL offers a substantial computational advantage over Monte Carlo methodologies (which must be based on static, not dynamic, decisions) and over DP. Furthermore, the MetaRL model has a capacity to handle larger state spaces where DP becomes computationally infeasible, underscoring its robustness and scalability. MetaRL's state space in this paper's implementation has 26 dimensions, which would absolutely be impossible to accommodate with DP. This RL approach provides fast, personalized wealth management strategies that are, on average, over 100 times faster than DP when implemented in inference mode for financial advising. Further, the MetaRL model is remarkably robust to changes in capital market environments, meaning that its accuracy when trained in one economic regime does not attenuate when applied to the other economic regimes it would inevitably encounter in any real-world situation. 

This paper shows how to extend the MetaRL methodology to GBWM problems that contain concurrent and partial goals. Also, it shows how to extend the MetaRL methodology to having stochastic inflation, which creates a four dimensional state space for DP that it cannot reasonably address, but a 27 dimensional state space for MetaRL that RL inference can address with ease. %Inflation increases can sharply diminish investor outcomes as goal costs rise with inflation and goals become less affordable. From our MetaRL analysis, we find that inflation volatility can diminish investor utility, though only when the mean reversion in inflation is low. Luckily, inflation mean reversion rates in the U.S. are high, mitigating the deleterious effect of inflation volatility. 
Future work will explore other extensions like this to the GBWM model, which were not feasible with DP, but now become explorable with MetaRL. Some of these are: enhancing GBWM with tax planning, regime-switching in the environment, optimal strategies in retirement for spending from multiple (taxable and tax-free) accounts, goal postponement \citep{bae_goal-based_2024}, pension planning, etc. %We believe it is also possible to handle GBWM alongside traditional wealth and utility maximization paradigms in a single unified framework using RL. 

%%%%%%%%%%%%%%%%%%%%%%%%%%%%%%%%%%%%%%%%%%%%%%%%%%%%%%%%%%%%%%%%
%% Appendices
%%%%%%%%%%%%%%%%%%%%%%%%%%%%%%%%%%%%%%%%%%%%%%%%%%%%%%%%%%%%%%%%
% \appendix

% \section{The first appendix}
% \label{sec:appendix1}
% This is an example of an appendix. 

% \noindent \textbf{Note:} Appendices appear before the references and are viewed as part of the ``main text'' and are subject to the 8--12 page limit, are peer reviewed, and can contain content central to the claims of the paper. 

% \section{The second appendix}
% \label{sec:appendix2}
% This is an example of a second appendix. If there is only a single section in the appendix, you may simply call it ``Appendix'' as follows:

% \section*{Appendix}
% % No label, since this can't be referenced meaningfully with \ref{}.
% This format should only be used if there is a single appendix (unlike in this document).

% \label{sec:ack}
% Use unnumbered third level headings for the acknowledgments. All acknowledgments, including those to funding agencies, go at the end of the paper. Only add this information once your submission is accepted and deanonymized. The acknowledgments do not count towards the 8--12 page limit.

%%%%%%%%%%%%%%%%%%%%%%%%%%%%%%%%%%%%%%%%%%%%%%%%%%%%%%%%%%%%%%%%
%% NOTE: THIS MARKS THE END OF THE "MAIN TEXT"
%%%%%%%%%%%%%%%%%%%%%%%%%%%%%%%%%%%%%%%%%%%%%%%%%%%%%%%%%%%%%%%%

%%%%%%%%%%%%%%%%%%%%%%%%%%%%%%%%%%%%%%%%%%%%%%%%%%%%%%%%%%%%%%%%
%% Bibliography
%%%%%%%%%%%%%%%%%%%%%%%%%%%%%%%%%%%%%%%%%%%%%%%%%%%%%%%%%%%%%%%%
\bibliography{Bib_for_Meta_RL}
\bibliographystyle{rlj}

%%%%%%%%%%%%%%%%%%%%%%%%%%%%%%%%%%%%%%%%%%%%%%%%%%%%%%%%%%%%%%%%
% AUTHOR: If your paper has no Appendixs, you may 
%         comment out the line below, which creates the title for
%         the Appendixs.
%%%%%%%%%%%%%%%%%%%%%%%%%%%%%%%%%%%%%%%%%%%%%%%%%%%%%%%%%%%%%%%%

% \beginSupplementaryMaterials

\clearpage

\appendix
\section*{Appendices}
\addcontentsline{toc}{section}{Appendices}
\renewcommand{\thesection}{\Alph{section}}

% Content that appears after the references are not part of the ``main text,'' have no page limits, are not necessarily reviewed, and should not contain any claims or material central to the paper. 
% %
% If your paper includes Appendixs, use the \begin{center}
%     {\tt {\textbackslash}beginSupplementaryMaterials} 
% \end{center}
% command as in this example, which produces the title and disclaimer above. 
% %
% If your paper does not include Appendixs, this command can be removed or commented out.

\section{RL Algorithm For At Most One All-Or-Nothing Goal In Each Time Step}\label{RLalgo}

The technical details of the RL algorithm are presented here, i.e., the RL environment, training and inference, and extensions. We restrict ourselves in this appendix to at most one goal in each time step, which the investor either completely fulfills or completely forgoes. In Appendix \ref{PCG}, we will extend our model to encompass the possibility of two or more goals in a time step (i.e., concurrent goals) and/or partial goals (for example, where an investor can choose some less nice vacations with reduced attained utilities but also reduced costs).

\subsection{The Meta-model's Environment}

We refer to our model as a ``meta-model'' \citep{beck_survey_2023} or, more specifically, our ``MetaRL'' model to distinguish it from RL approaches that are trained to solve a single task. To create our MetaRL model, we use pre-trained RL models to handle scenarios that are arbitrary in terms of (1) the time horizon, (2) the initial wealth and the times and amounts of any subsequent wealth infusions, and (3) the number of goals and, for each goal, the time it can be purchased, the cost to purchase it at that time, and the utility (representing the goal's importance to the investor) gained by the investor if the goal is purchased. We may think of this approach as being similar to training for zero-shot learning and inference \citep{pourpanah2022review}.

\subsection{Actions And The Environment Over A Time Step}
\label{sec:env_wealth_transition}
We describe the environment for a single time step between time $t$ and time $t+1$, using the notation for scenario parameters introduced in Subsection \ref{sec:param}. We define time $t^-$ to be at time $t$, just before the goal-taking choice is made, and time $t^+$ to be at time $t$, but just after the goal-taking choice is made. At time $t^-$, the goal-taking agent (GoalAgent) produces an action $a_g(t) \in [0,1]$. If its value is below a threshold value $a_{\mathrm{thresh}}$ or if there is not sufficient money (that is, $W(t^-)<C(t)$), the goal is not taken, in which case the wealth is unchanged ($W(t^+) = W(t^-)$) and there is no attained utility ($U_{\mathrm{attained}}(t)=0$). On the other hand if $a_g(t) \ge a_{\mathrm{thresh}}$ and $W(t^-)\ge C(t)$, then the goal is taken, in which case, we subtract the cost of the goal from the wealth and set $U_{\mathrm{attained}}(t)=U(t)$, the utility associated to the goal. That is, in either the case of not taking the goal or taking the goal, if we define the decision $g(t) \in \{0,1\}$ by
\begin{equation}\label{gdef}
g(t)= \mathbbm{1}_{a_g(t)\geq a_{\mathrm{thresh}}} \cdot \mathbbm{1}_{W(t^-)\geq C(t)},  
\end{equation} 
we have
\begin{eqnarray}
    W(t^+) &=& W(t^-) - g(t)\cdot C(t) 
    \label{eq:goaltaking}\\
    U_{\mathrm{attained}}(t)&=&g(t)\cdot U(t).
    \label{eq:utilitytaking}
\end{eqnarray}
We note that we set $a_{\mathrm{thresh}} = 0.5$ for the experiments in this paper, since 0.5 is the midpoint of $[0,1]$, the range for $a_g(t)$.

The resulting wealth, $W(t^+)$, is used to compute the state input for the investment portfolio agent (PortfolioAgent).  PortfolioAgent returns the action $a_p(t)\in[0,1]$, which defines $p(t)$, the investment portfolio decision, by setting $p(t)$ equal to the value of the integer $p$ where
\begin{equation}\label{pdef}
\frac{p}{P}\le a_p(t)<\frac{p+1}{P}
\end{equation}
(and setting $p=P-1$ if $a_p(t)=1$).
So, for example, if $p$ increasing corresponds to progressively aggressive portfolios (potentially along the efficient frontier, although that doesn't need to be the case), then the magnitude of $a_p(t)$ corresponds to how aggressive the chosen investment portfolio is. We then assume that the wealth between time $t$ and time $t+1$ will evolve via geometric Brownian motion (GBM) with the chosen portfolio $p$. That is,
\begin{equation}
W((t+1)^-) = W(t^+)\cdot \mathrm{exp}\left[ {(\mu_p-\frac{1}{2}\sigma_p^2)\cdot h + \sigma_p \cdot Z \cdot \sqrt{h}}\right] + I(t+1).
\label{eq:portgbm}
\end{equation}
Note that our model here assumes that the timing and magnitude of the infusions, $I(t)$, can either be predicted or known in advance. Of course, unlike the initial wealth, which can be applied to any goal, an infusion at time $t$ can only be applied to goals available at time $t$ or later.

The objective of both the agents combined is to maximize $E\left[\sum_{\tau=t}^T U_{\mathrm{attained}}(\tau) \right]$, the expected value of the utility accrued over the remaining time horizon of the problem. Since the evolution equations \eqref{eq:goaltaking}, \eqref{eq:utilitytaking}, and \eqref{eq:portgbm}, as well as the optimal objective, are dependent only on the time, $t$, the current wealth, $W(t)$, and the action choices, we conclude that the problem is Markovian, and hence suitable for a RL formulation.

While the goal-taking and portfolio agents interact independently with the environment (including state observation sampling, storing trajectory memories, and gradient propagation during training), their rewards are interdependent since they interact iteratively with the environment.

\subsection{States} \label{sec:state}

In principle, the scenario parameters introduced in Subsection \ref{sec:param} provide sufficient information to compute $a_g(t)$ and $a_p(t)$. However, we require some manipulation of these parameters in order to obtain more effective state variables for the MetaRL agents. The objectives of this manipulation are, (i) to define a constant-dimension input for the MetaRL agent regardless of the time $t$ being considered and the time horizon $T$, and (ii) to provide rich information that helps the MetaRL agent identify patterns during training. This subsection describes these more effective state variables. 

We look to have state variables that are unitless and normalized to either only take values between 0 and 1 or have most of their interesting behavior occur between 0 and 1. This makes the gradient ascent algorithm in PPO more effective, since gradient ascent is well known to work poorly with independent variables that have significantly different scales.

The first three state variables are the scalars $t_\mathrm{norm}$, $W_{\min}$, and $W_{\max}$, which are unitless, normalized versions of the time, $t$, and wealth, $W(t)$. The definition of $t_\mathrm{norm}$ is simply
\begin{equation}
t_\mathrm{norm} = \frac{t}{T}.    
\end{equation}

The important aspect about the wealth at time $t$ is its ability to obtain current (time $=t$) and future (time $>t$) goals, so we look to normalize the wealth in relation to the sum of the costs of all current and future goals. To do this, we must approximate how to discount the costs of future goals to time $t$ dollars, which $W(t)$ is obviously expressed in. Equation \eqref{eq:portgbm}, the GBM equation, gives the expression for discounting if we set the infusions to zero. This equation depends on the future portfolios, $p$, that are chosen, which can shift over time, and the future stochastic behavior, given by $Z$. The algorithm \textsc{DiscountSum}$({\bf C}[t:],p,z)$, given in Algorithm \ref{alg:discount}, uses equation \eqref{eq:portgbm} to compute the discounted sum of current and future goal costs, assuming a given, fixed portfolio, $p$, for all times $\ge t$ and a given, fixed value $z$ in place of $Z$ for all times $\ge t$.

For the purposes of normalizing the wealth, we will approximate discounting the future goal costs in two different ways by using \textsc{DiscountSum}$({\bf C}[t:],p,z)$ with two different $(p,z)$ pairs. For the first pair, we set $p$ to be 0 (the most conservative investment portfolio) and $z=-1$ (a pessimistic scenario). For the second pair, we set $p=P-1$ (the most aggressive investment portfolio) and $z=1$ (an optimistic scenario). This gives us $W_{\min}$ and $W_{\max}$, our two state variables for the normalized wealth:
\begin{eqnarray}
W_{\min} & = & \frac{W(t)}{\textsc{DiscountSum}({\bf C}[t:],p=0,z=-1)} \\
W_{\max} & = & \frac{W(t)}{\textsc{DiscountSum}({\bf C}[t:],p=P-1,z=1)}. 
\end{eqnarray}

The second set of three state variables are the vectors ${\bf{U}}_\mathrm{agg}$, ${\bf{C}}_{\min}$, and ${\bf{C}}_{\max}$, which describe the relative timing of the utilities and costs of goals for times $\ge t$. It matters if we are looking to attain a goal that occurs one year from now versus, say, 23 years from now. These state variable vectors look to contain the key aspects of this timing information, while retaining a key property for RL computation of having a fixed length, regardless of the time $t$ at which they are computed or the time horizon $T$. 

\begin{algorithm}[h!]
{\footnotesize 
\caption{Discounting future goal costs and returning each of them (\textsc{DiscountVec}) or their sum (\textsc{DiscountSum}).}
\label{alg:discount}
\begin{algorithmic}
\Procedure{DiscountVec}{${\bf{C}}[t:]$, $p$, $z$}
    \State ${\bf{C}}_\mathrm{disc} \gets$ initialize vector of zeros of same length as ${\bf{C}}[t:]$, which is $T-t+1$
    \For{$\tau$ in $[0:$ length$({\bf{C}}[t:])]$}
        \State ${{C}}_\mathrm{disc}(\tau) = {{C}}(\tau)\;\mathrm{exp}\left[ {-(\mu_p-\frac{1}{2}\sigma_p^2)h\tau - \sigma_p  z \sqrt{h\tau}}\right]$
    \EndFor
    \State \textbf{return} ${\bf{C}}_\mathrm{disc}$
\EndProcedure
\\
\Procedure{DiscountSum}{${\bf{C}}[t:]$, $p$, $z$}
    \State \textbf{return} sum$\big(\textsc{DiscountVec}({\bf{C}}[t:], p, z) \big)$
\EndProcedure
\end{algorithmic}
}
\end{algorithm}

We begin with the utilities. At first, one might simply think to use the vector ${\bf U}[t:]$, which contains the utility of each current (time $t$) and future goal, and then divide this vector by its sum to normalize its components to be between 0 and 1. However, because the length of ${\bf U}[t:]$ is $T-t+1$, which varies depending on $t$ and $T$, and we need our state variable utility vector to have the same length regardless of $t$ and $T$, we instead look to aggregate ${\bf U}[t:]$ into a fixed number, $K$, of time blocks. As an example that we will use in all of our experiments, we consider the following list {\bf L} that contains $K=7$ aggregated time blocks: ${\bf{L}}=[[0], [1], [2], [3], [4,5], [6,7,8,9], [10:]]$. The penultimate time block for this ${\bf L}$ specifies aggregating all the utilities from goals occurring at times $t+6$ through $t+9$, while the final time block specifies aggregating all the utilities from goals occurring at time $t+10$ or later. 

To define ${\bf U}_\mathrm{agg}$, our state variable utility vector, we apply the aggregation algorithm {\sc Aggregate}$({\bf V},{\bf L})$, defined in Algorithm \ref{alg:aggregate}, to the vector ${\bf V}={\bf U}[t:]$ and then divide by the sum of the components in ${\bf U}[t:]$: 
\begin{equation}
{\bf{U}}_\mathrm{agg}[t:] = \frac{\textsc{Aggregate}({\bf U}[t:], {\bf L})}{\sum_{\tau=t}^T U(\tau)}.    
\end{equation}
Note that $\textsc{Aggregate}({\bf U}[t:], {\bf L})$ is a vector of length $K=7$ where, for example, the final component contains the sum of the utilities from the goals that occur at time $t+10$ or later.

Our approach to the state variables for the goals' costs is parallel to our approach for ${\bf{U}}_\mathrm{agg}$, the goals' utilities, but since we're looking to compare relative costs, we need to discount the costs to time $t$ dollars for a fair comparison. To do this we use the same approach to discounting (along with the same two $(p,z)$ pairs) that we used to define $W_{\min}$ and $W_{\max}$ to obtain
\begin{eqnarray}
{\bf{C}}_{\min}[t:] &=& \frac{\textsc{Aggregate}\big(\textsc{DiscountVec}({\bf C}[t:],p=P-1,z=1), {\bf L}\big)}{\textsc{DiscountSum}({\bf C}[t:],p=P-1,z=1)} \\
{\bf{C}}_{\max}[t:] &=& \frac{\textsc{Aggregate}\big(\textsc{DiscountVec}({\bf C}[t:],p=0,z=-1), {\bf L}\big)}{\textsc{DiscountSum}({\bf C}[t:],p=0,z=-1)},
\end{eqnarray}
where $\textsc{DiscountVec}({\bf C}[t:],p,z)$ is defined in Algorithm \ref{alg:discount}. 
We note that because the discounted sum is in the denominator when defining $W_{\min}$ and $W_{\max}$, while the key discounting is in the numerator when defining ${\bf{C}}_{\min}$ and ${\bf{C}}_{\max}$, the case where $(p,z) = (0,-1)$ corresponds to $W_{\min}$ and $C_{\max}$, while the case where $(p,z) = (P-1,1)$ corresponds to $W_{\max}$ and $C_{\min}$.

\begin{algorithm}[h!]
{\footnotesize
\caption{Aggregating {\bf V}, which can represent a vector of goal utilities or a vector of goal costs, over the $K$ aggregated current and future time blocks given in {\bf L}.}
\label{alg:aggregate}
\begin{algorithmic}
\Procedure{Aggregate}{Vector ${\bf{V}}$, List of lists ${\bf{L}}$}
    \State $K \gets$ length of ${\bf{L}}$ 
    \State ${\bf{A}} \gets$ vector of zeros of length $K$ 
    \For{$\ell$ in $[0:K]$}
        \State {\bf currentlist} $\gets {\bf{L}}(\ell)$
        \For{$i$ in $[0:$ length$(\mathrm{\bf currentlist})]$}
        \State ${{A}}(\ell) \gets A(\ell) + V(\mathrm{\bf currentlist}(i))$
        \EndFor
    \EndFor
    \State \textbf{return} ${\bf{A}}$
\EndProcedure
\end{algorithmic}
}
\end{algorithm}

The final two state variables are the scalars $\gsim$ and $\psim$.  These are key state variables with $\gsim$ aimed to be particularly useful for the GoalAgent and $\psim$ for the PortfolioAgent. These two state variables provide the MetaRL agents with richer information about the dynamics of the environment by running a handful of deterministic forward simulations, using some considerable approximations for the future dynamics. The procedure is inspired by literature on Neural Monte-Carlo Tree Search \citep{silver_mastering_2016}, but in simplified form.

More specifically, $\gsim$ and $\psim$ are used as indicator variables to provide rough guidance to the RL policy. They are more important early in the training, when long time horizons corresponding to several portfolio choices and goal choices threaten to derail value estimates. Towards the latter part of training, the policy learns when to deviate from these indicator variables (for example, taking low-confidence goals against the indicated recommendation when future opportunities are scarce). This is discussed further in Appendix \ref{sec:Rew}.

To compute either $\gsim$ or $\psim$, we start by selecting a positive integer $n$ and then determining a list of the $n$ values of the standard normal $Z$ that are the midpoints of a partition of the cumulative standard normal distribution into $n$ equal parts. That is, the list's components, $z_i$ where $i=0,1,...,n-1$, are defined so that $P(Z<z_i) = \frac{i+\frac{1}{2}}{n}.$ Since we find that $n=11$ is sufficiently large for our results, we use this in all our experiments. 

We then calculate $\textsc{DiscountVec}({\bf C}[t:],p,z_i)$ and $\textsc{DiscountVec}({\bf I}[t:],p,z_i)$ for each investment portfolio $p$ and each $z_i$. This gives the goal costs and infusions in time $=t$ dollars, given the investment portfolio is fixed to be $p$ and $Z=z_i$ in equation \eqref{eq:portgbm} for all times $\ge t$.  For each $(p,z_i)$ pair, we can then take the wealth, $W(t^-)$, and the discounted infusions, which are all in time $t$ dollars, and use them to purchase goals in order of decreasing utility (using chronological order for goals with the same utility). In purchasing these goals, we pay for the discounted goal prices $\textsc{DiscountVec}({\bf C}[t:],p,z_i)$ by first applying any available discounted infusions $\textsc{DiscountVec}({\bf I}[t:],p,z_i)$ (using reverse chronological order, meaning using the most recently available infusion that can be applied to the goal) and then, if all the available infusions are drained, applying $W(t^-)$. We stop when the discounted infusions and $W(t^-)$ remaining are too small to purchase any more goals or all the goals have been purchased. This gives the accumulated utility as a function of $p$ and $z_i$. For each fixed $p$, we then take the average of the accumulated utilities over the $n$ values of $z_i$ to give a loose approximation of the expected accumulated utility sums for each investment portfolio $p$. 

To determine $\gsim$, we use the method just described to determine the approximated expected utility sums (for each value of $p$) for two different sets of goals: one where we first take the current (time $=t$) goal, if it is possible, before taking the future goals in descending order of their utilities as described above, the other where we remove the current goal completely before taking the future goals, again as described above. Since we are looking to optimize these two approximated expected utility sums, we only retain the largest approximated expected utility sum over the $P$ values of $p$.\footnote{We note that this is a significant approximation because we have fixed $p$ for all current and future times, when our final results will allow the investment portfolio $p$ to vary with time. Similarly, our expected accumulated utility sums only used fixed values $Z=z_i$ as a significant approximation for potentially complex Brownian motion. These approximations work well in practice, as we will see, and they can be computed deterministically and quickly, which is key for our state variable calculations.} This approximates using the best (fixed) portfolio choice, which may, of course, correspond to different values of $p$ for the two approximated expected utility sums. We then subtract the second optimized approximated expected utility sum (which skips the time $t$ goal) from the first sum (which gives the time $t$ goal first priority), then divide this difference by the first sum to normalize it, and finally define $\gsim$ to be the logistic function of this normalized difference, which keeps $\gsim \in [0,1]$. Note that the larger $\gsim$ is, the more evidence there is to take the goal at time $t$, which is exactly what GoalAgent is looking to decide. This process that we have just described for computing $\gsim$ is summarized in Algorithm \ref{alg:gsim}. Given the nature of $\gsim$, we may consider it akin to a binary classifier, choosing between one of two goal choices.

To determine $\psim$, we use a similar process. As before, we compute the approximated expected utility sums for each $p$, but we no longer give first preference to the goal at time $t$, nor do we remove it. We then find the $p$ that gives the highest approximated expected utility sum, and normalize by dividing it by $P-1$ to determine the value for $\psim \in [0,1]$. If the investment portfolios get more aggressive as $p$ increases, as we will have in our experiments where the investment portfolios are chosen over an efficient frontier, the higher $\psim$ is, the more evidence to choose a more aggressive investment portfolio. This process for computing $\psim$ is summarized in Algorithm \ref{alg:psim}. Unlike the binary classifier nature of $\gsim$, the value of $\psim$ is used as a regressor that chooses one portfolio from an ordered set of portfolios.

\begin{algorithm}[h!]
{\footnotesize
\caption{Forward simulation to compute $\gsim$}
\label{alg:gsim}
\begin{algorithmic}
\Procedure{GSim}{$W(t^-)$, ${\bf{C}}[t:]$, ${\bf{U}}[t:]$, $\boldsymbol{\mu}$, $\boldsymbol{\sigma}$, $n$}
    \State {\bf listZ} $\gets$ the $n$ midpoint $Z$ values after partitioning Z's CDF into $n$ equal sections 
    \State {\bf sortU} $\gets$ components of ${\bf{U}}[t:]$ in descending order (with ties in chronological order)
    \State Force the goal at time $=t$ to be the first entry in {\bf sortU}
    \State ${\bf{E}}_\mathrm{take} \gets$ vector of zeros of length $P$
    \For{$p$ in $[0,\ldots,P-1]$}
        \State ${\bf{C}}_\mathrm{disc}, {\bf{I}}_\mathrm{disc},$ and ${\bf U}_\mathrm{tot} \gets$ vectors of zeros, each of length $n$
        \For{$z$ in {\bf listZ}}
            \State ${\bf{C}}_\mathrm{disc} \gets$ \textsc{DiscountVec}({\bf C}[t:],p,z)
            \State ${\bf{I}}_\mathrm{disc} \gets$ \textsc{DiscountVec}({\bf I}[t:],p,z)
            \State ${\bf sortC} \gets {\bf{C}}_\mathrm{disc}$ reordered to be in the same goal order as {\bf sortU}
            \State ${\bf U}_\mathrm{tot}[z] \gets$ accrued utility by spending ${\bf{I}}_\mathrm{disc}$ and $W(t^-)$ in the order of goals in ${\bf sortU}$,  
            \State \hspace{3 em} paying amounts specified in ${\bf sortC}$, until no more goals can be attained
        \EndFor
        \State ${\bf{E}}_\mathrm{take}[p] \gets$ average of ${\bf U}_\mathrm{tot}$ over its $n$ components
    \EndFor
    \State Remove first goal from {\bf sortU}
    \State ${\bf{E}}_\mathrm{skip} \gets$ vector of zeros of length $P$
    \For{$p$ in $[0,\ldots,P-1]$}
        \State ${\bf{C}}_\mathrm{disc}, {\bf{I}}_\mathrm{disc},$ and ${\bf U}_\mathrm{tot} \gets$ vectors of zeros, each of length $n$
        \For{$z$ in {\bf listZ}}
            \State ${\bf{C}}_\mathrm{disc} \gets$ \textsc{DiscountVec}({\bf C}[t:],p,z)
            \State ${\bf{I}}_\mathrm{disc} \gets$ \textsc{DiscountVec}({\bf I}[t:],p,z)
            \State ${\bf sortC} \gets {\bf{C}}_\mathrm{disc}$ reordered to be in the same goal order as {\bf sortU}
            \State ${\bf U}_\mathrm{tot}[z] \gets$ accrued utility by spending ${\bf{I}}_\mathrm{disc}$ and $W(t^-)$ in the order of goals in ${\bf sortU}$,  
            \State \hspace{3 em} paying amounts specified in ${\bf sortC}$, until no more goals can be attained
        \EndFor
        \State ${\bf{E}}_\mathrm{skip}[p] \gets$ average of ${\bf U}_\mathrm{tot}$ over its $n$ components
    \EndFor
    \State $\max_\mathrm{take} \gets \max({\bf{E}}_\mathrm{take})$ and $\max_\mathrm{skip} \gets \max({\bf{E}}_\mathrm{skip})$
    \State \textbf{return} logistic$((\max_\mathrm{take}-\max_\mathrm{skip})/\max_\mathrm{take})$
\EndProcedure
\end{algorithmic}
}
\end{algorithm}

\begin{algorithm}[h!]
{\footnotesize
\caption{Forward simulation to compute $\psim$}
\label{alg:psim}
\begin{algorithmic}
\Procedure{PSim}{$W(t^-)$, ${\bf{C}}[t:]$, ${\bf{U}}[t:]$, $\boldsymbol{\mu}$, $\boldsymbol{\sigma}$, $n$}
    \State {\bf listZ} $\gets$ the $n$ midpoint $Z$ values after partitioning Z's CDF into $n$ equal sections 
    \State {\bf sortU} $\gets$ components of ${\bf{U}}[t:]$ in descending order (with ties in chronological order)
    \State ${\bf{E}} \gets$ vector of zeros of length $P$
    \For{$p$ in $[0,\ldots,P-1]$}
        \State ${\bf{C}}_\mathrm{disc}, {\bf{I}}_\mathrm{disc},$ and ${\bf U}_\mathrm{tot} \gets$ vectors of zeros, each of length $n$
        \For{$z$ in {\bf listZ}}
            \State ${\bf{C}}_\mathrm{disc} \gets$ \textsc{DiscountVec}({\bf C}[t:],p,z)
            \State ${\bf{I}}_\mathrm{disc} \gets$ \textsc{DiscountVec}({\bf I}[t:],p,z)
            \State ${\bf sortC} \gets {\bf{C}}_\mathrm{disc}$ reordered to be in the same goal order as {\bf sortU}
            \State ${\bf U}_\mathrm{tot}[z] \gets$ accrued utility by spending ${\bf{I}}_\mathrm{disc}$ and $W(t^-)$ in the order of goals in ${\bf sortU}$,  
            \State \hspace{3 em} paying amounts specified in ${\bf sortC}$, until no more goals can be attained
        \EndFor
        \State ${\bf{E}}[p] \gets$ average of ${\bf U}_\mathrm{tot}$ over its $n$ components
    \EndFor
    \State \textbf{return} $\arg\hspace{-.15 em}\max ({\bf{E}})/(P-1)$
\EndProcedure
\end{algorithmic}
}
\end{algorithm}

Experiments with and without $\gsim$ and $\psim$ show that their role during policy deployment is quite significant. In fact, experiments also show that the results of decisions based {\it solely} on $\gsim$ and $\psim$ are fairly close to those taken by using the full list of our state variables. This full list of state variables is summarized in Table \ref{tab:observations}.
\begin{table}[h!]
\centering
\caption{\small Summary of state variable inputs for the two action producing agents, GoalAgent and PortfolioAgent. For all experiments in this paper, we use a value of $K=7$ aggregated current and future time blocks, which are given by the list ${\bf{L}}=[[0], [1], [2], [3], [4,5], [6,7,8,9], [10:]]$, where numbers represent time steps in the future.}
\label{tab:observations}
{\small 
\begin{tabular}{|p{1.5cm}|p{2.3cm}|p{5.8cm}|}
\hline
Symbol & Length & Explanation \\ \hline
$t_\mathrm{norm}$ & 1 (float) & Time step normalized by the final time step horizon $T$  \\
$W_\mathrm{\min}$ & 1 (float) & Current wealth normalized by the sum of future goal costs discounted using a pessimistic, conservative scenario \\
$W_\mathrm{\max}$ & 1 (float) & Current wealth normalized by the sum of future goal costs discounted using an optimistic, aggressive scenario \\
${\bf{U}}_\mathrm{agg}$ & $K$ (float) & Vector of utilities aggregated into $K$ time blocks and normalized by their sum\\ 
${\bf{C}}_{\min}$ & $K$ (float) & Vector of optimistically, aggressively discounted goal costs aggregated into $K$ time blocks and normalized by their sum \\
${\bf{C}}_{\max}$ & $K$ (float) & Vector of pessimistically, conservatively discounted goal costs aggregated into $K$ time blocks and normalized by their sum \\
$\gsim$ & 1 (float) & Forward-simulation-based evidence for taking the current goal\\
$\psim$ & 1 (float) & Forward-simulation-based evidence for selecting a more aggressive investment portfolio at the current time\\
\hline
\end{tabular}
}
\end{table}

% \subsubsection{Actions}

% The state inputs as described above are sent to the relevant agent (GoalAgent or PortfolioAgent), which return scalar actions $a_g(t)$ and $a_p(t)$ respectively in the range $[0,1]$. For the GoalAgent, the action is interpreted as the confidence regarding taking the current goal, and is therefore rounded off to $0$ or $1$ for implementation. For the PortfolioAgent, the action is interpreted as a `level of aggression' of the chosen investment portfolio, with $0$ corresponding to the most conservative and $1$ corresponding to the most aggressive. This action is mapped proportionally to an integer value in $\{0,\ldots,P-1\}$ in order to pick the investment portfolio to invest in.

\subsection{Rewards}
\label{sec:Rew}

We define two types of rewards associated with each time step for training the two MetaRL agents. The first type of time step reward, the extrinsic reward, $r_e(t)$, corresponds to the underlying objective function of the problem. We define this at all time steps, except the final time step at $T$, by
\begin{equation}
    r_e(t) = \frac{g(t) \cdot U(t)}{\sum_{\tau=0}^T U(\tau)} \text{ for } t<T.
    \label{eq:ex_rew_1}
\end{equation}
Recall from the definition of $g(t)$ in equation \eqref{gdef}, this means the reward is accrued if and only if (i) sufficient wealth is available to take the goal, and (ii) the GoalAgent action, $a_g(t)$, is higher than $a_{\mathrm{thresh}}$, where we set $a_{\mathrm{thresh}}=0.5$ for the experiments in this paper.  

At the final time step $t=T$, we augment the above definition by adding a small reward to the agent for the amount of money remaining if it is not enough to take the final goal. This signals to the MetaRL algorithm that finishing with more money is better than less money, which creates an otherwise missing push towards attaining the final goal as the MetaRL model evolves over epochs. Specifically, we define the extrinsic reward at the final step $T$ by
\begin{equation}
    r_e(T) = \frac{\mathbbm{1}_{a_g(T)\geq a_{\mathrm{thresh}}} \cdot U(T)}{\sum_{\tau=0}^T U(\tau)} \left[ \mathbbm{1}_{W(T^-)\geq C(T)} + \mathbbm{1}_{W(T^-)< C(T)} \cdot \frac{1}{4} \left(\frac{W(T^-)}{C(T)}\right) \right].
    \label{eq:ex_rew_2}
\end{equation}

Even with this modification at $t=T$, the extrinsic reward provides a relatively weak signal to the MetaRL agents because (i) the rewards are driven by a number of factors, such as the initial wealth and the stochasticity in geometric Brownian motion, that are disconnected from the MetaRL agents' decisions, (ii) the rewards are often delayed a number of time steps after MetaRL agent decisions are made, and (iii) the dual agent architecture makes it difficult to accomplish credit assignment \citep{zhou2020learning}.  We therefore provide a second reward, called the intrinsic reward, at each time step for each of the two MetaRL agents, following literature on intrinsic motivation in reinforcement learning \citep{barto2013intrinsic}. The intrinsic rewards are penalties (that is, negative rewards) for actions being far away from their corresponding forward simulated state variables. 

More specifically, for GoalAgent and PortfolioAgent, the intrinsic rewards are respectively given by
\begin{eqnarray}
    r_{i,g}(t) &=& -\frac{1}{2}\rho\;
    %\frac{\sum_{\tau=t}^T \gamma^{(\tau -t)}U(\tau)}{\sum_{\tau=t}^T U(\tau)}\;
    |\gsim-a_g(t)| \mbox{ and} \label{eq:in_rew_g}\\
    r_{i,p}(t) &=& -\frac{1}{2}\rho\;
    % \frac{\sum_{\tau=t+1}^T \gamma^{(\tau -t)}U(\tau)}{\sum_{\tau=t+1}^T U(\tau)}\;
    |\psim-a_p(t)|,
    \label{eq:in_rew_p}
\end{eqnarray}
where the coefficient $\rho$ is a scalar multiple that we anneal from $1$ to $0.25$ over the course of the MetaRL training to emphasize the extrinsic reward more by the end of the training. 
With $r_e(t), r_{i,g}(t)$, and $r_{i,p}(t)$ defined, we define $R_g(t)$ and $R_p(t)$, the total rewards used at time $t$ for respectively training GoalAgent and PortfolioAgent, by\footnote{Note that for the sums over time steps in the definitions of both $R_g(t)$ and $R_p(t)$, we have implicitly used a discount factor of 1 (i.e., no discounting) to reflect the fact that the importance of a goal (i.e., its utility) is not diminished by when it occurs in the future.}
\begin{eqnarray*}
    R_g(t) &=& \sum_{\tau=t}^T \big(r_e(\tau)+r_{i,g}(\tau)\big) \mbox{ and}\\
    R_p(t) &=& \sum_{\tau = t+1}^T r_e(\tau) + \sum_{\tau=t}^T r_{i,p}(\tau),
\end{eqnarray*}
noting that the value of $r_e(t)$, the extrinsic reward at time $t$, is excluded from the definition of $R_p(t)$ because PortfolioAgent is called {\it after} the goal-taking decision has been executed at time $t$.

This completes the description of the formulation of the MetaRL environment. We next discuss implementation details of this formulation.

\section{MetaRL Model Training And Inference}
\label{sec:training_inference} 

\subsection{Network Architecture And Hyperparameters}
\label{nn_architecture}

RL formulates the decision-making at each time step to consist simultaneously of the goal-taking choice and the investment portfolio selection. We note that the goal-taking choice at the beginning of any time step directly affects the subsequent investment portfolio selection during that time step, because taking a goal affects the available remaining wealth, which affects the optimal investment portfolio choice. Therefore, the RL methodology depicted in Figure \ref{fig:flowchart} splits the decision into two steps for goal-taking and investment portfolio selection (shown in the ``Agents'' section of Figure \ref{fig:flowchart}). 
% \begin{figure}[ht!]
%     \centering
%     \includegraphics[width=0.99\textwidth]{RL_flowchart.png}
%     \caption{ \small Logical flow of the RL methodology for a single ``episode,'' during which we have time steps $t=0,1,...,T$. Typically, RL runs through a large number of such episodes during training in order to generalize learning to arbitrary problems. The same logical flow is followed during inference on any new problem. }
%     \label{fig:flowchart}
% \end{figure}

Because the Proximal Policy Optimization (PPO) algorithm (see \cite{schulman_trust_2015, schulman_proximal_2017, haarnoja_soft_2018}) is based on the actor-critic architecture, it contains two neural networks (actor and critic) for GoalAgent and also for PortfolioAgent. The dimension of the input into the actor and critic networks is the dimension of the state space, which, from Table \ref{tab:observations}, is $5+3K = 26$, when $K=7$, as we assume in our experiments. While GoalAgent and PortfolioAgent are trained as if they are independent PPO agents, the logical flow is similar to Hierarchical RL \citep{barto2003recent}.

The output for both actors and both critics are scalars, corresponding to $a_g$ and $a_p$ for the two actors, and RL value functions for the two critics. The three intermediate layers for the actor networks have 256, then 64, and finally 16 nodes, while the two for the critic networks have 64 and then 16 nodes. These numbers of nodes and layers, as well as the other hyperparameter values given in the next paragraph, were determined on the basis of an experimental hyperparameter sweep. We do not use any special initializers for the parameters of the networks. 

Both the actor and critic networks use the \texttt{tanh} activation function for their intermediate layers. The critics have a \texttt{linear} output neuron to ensure the full range of predictions of returns-to-go (that is, RL value function predictions), while the actors have a \texttt{sigmoid} output neuron to ensure the output is in the range $[0,1]$. For hyperparameter values, we used a learning rate of $10^{-4}$, a clip parameter of 0.2, and a discount rate of 1.0 (meaning no discounting, since future utility values are not of lesser importance than current utility values). 

We trained the MetaRL model using 1000 epochs, where each epoch runs for 500 episodes. All networks were updated at the end of each epoch. Training the MetaRL model using 1000 epochs took approximately 4 hours.

\subsection{Training Curriculum}
\label{sec:curriculum}

In order to maximize the generalizability of the agents, we provided the MetaRL algorithm with a diverse set of scenarios during training. These scenarios were generated using the procedure \textsc{GenerateScenario} described in Algorithm \ref{alg:genprob}, which is designed to explore a wide range of time horizons and numbers of goals, along with varying the goals' timing, costs, and utilities. A fresh scenario is generated at the start of every epoch, which is solved by a set of 500 random episodes. The randomness between these 500 episodes has three sources: (1) The values generated for $Z$, the standard normal random variables that determine the portfolio returns in equation \eqref{eq:portgbm}, (2) the randomness used to determine exploration for the action variables $a_g$ and $a_p$, (3) a small randomization in the initial wealth, which is selected from a uniform distribution between 80\% and 120\% of $W(0)$, the initial wealth for the epoch given by the \textsc{GenerateScenario} procedure. The collected experience is used for training the critic and actor networks at the end of each epoch. Note that the intrinsic reward from \eqref{eq:in_rew_g} and \eqref{eq:in_rew_p} is being annealed over the course of training, as explained in Appendix \ref{sec:Rew}. 

The above process is done in parallel using five different seeds for the randomness.\footnote{We use seeds 0, 15, 722, 1021, 5069 in this paper for all reported results.} That is, in some sense we are running the MetaRL model five times, but the scenarios of all five are the same in each epoch, only the randomness for the 500 episodes within each epoch is different due to the different seeds. The actions produced during inference are determined by the median action produced by the five models.

\begin{figure}[ht!]
\centering
\includegraphics[width=\linewidth]{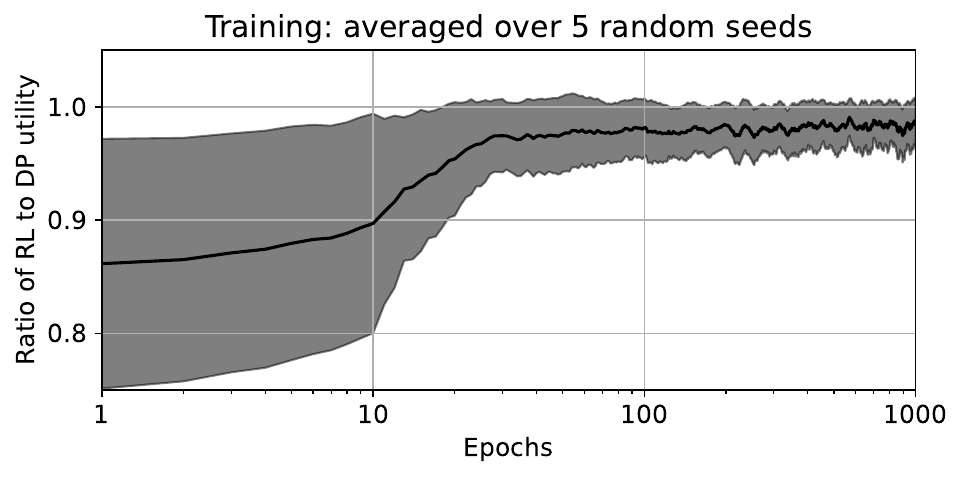}
\caption{\small Training efficiency of the RL algorithm over the course of training (using the RL-Efficiency for each of the 1000 epochs' scenarios). While the training episodes each have randomised initial wealth, the quantities plotted here are for the baseline values of initial wealth. The black line represents the average of the five RL-Efficiencies determined using five different seeds for randomness. The shaded region corresponds to being one standard deviation away from the average, where the standard deviation is computed from the five RL-Efficiencies.}
\label{fig:training}
\end{figure}

Figure \ref{fig:training} was compiled to test whether 1000 epochs is sufficient. It shows that it is more than enough, with a large fraction of the algorithm's learning being completed within the first 100 epochs. The figure uses the ``RL-Efficiency'' determined from all five models. The RL-Efficiency is defined in Subsection \ref{sec:EUA}, except that the denominator here is the DP value function (also discussed in that subsection). The bold black line in the graph represents the average RL-Efficiency from the five models. The gray area corresponds to adding or subtracting a standard deviation of the five RL-Efficiencies from the average RL-Efficiency.
\begin{algorithm}[tb]
{\footnotesize
\caption{Generating a training scenario}
\label{alg:genprob}
\begin{algorithmic}
\Procedure{GenerateScenario}{$\boldsymbol{\mu}$, $\boldsymbol{\sigma}$}
    \State Generate $T$ (uniformly) randomly from $\{5,6,...,50\}$
    \State ${\bf{C}}, {\bf{U}} \gets$ arrays of zeros of length $T+1$
    \State Generate $N_G$, the number of goals, from the distribution $p(1)=0.22, p(2)=0.15, p(3)=0.12, p(4)=$
    \State \hspace{0.6 em}$0.10, p(5)=0.06, p(6)=0.05, p(7)=0.04, p(8)=0.03, p(9)=0.02, p(10)=0.01, p(T)=0.20$
    \If{$N_G>T$}
        \State $N_G \gets T$
    \EndIf
    \State One of the $N_G$ goal times is $T$. Randomly choose the other $N_G-1$ goal times from $\{1,2,...,T-1\}$
    \For{each non-zero goal time $t\in[1,T]$}
        \State Draw numbers $u_1$ and $u_2$ from a uniform distribution on $[0,1]$
        \State $C(t) \gets 100\cdot u_1\cdot 1.03^t$
        \State $U(t) \gets 0.3\cdot C(t)/1.03^t + 25\cdot u_2$
    \EndFor
    \State Choose the initial wealth $W(0)$ from a uniform distribution between  \textsc{DiscountSum}$({\bf C}[:], P-1, 2)$ and 
    \State \hspace{2 em} \textsc{DiscountSum}$({\bf C}[:], 0, -2)$, where  \textsc{DiscountSum} is definied in Algorithm \ref{alg:discount}
    \State \textbf{return} ${\bf C}[1:], {\bf U}[1:], W(0)$
\EndProcedure
\end{algorithmic}
}
\end{algorithm}

\subsection{Implementation Details}
\label{sec:app_implementation_details}

The RL algorithm (and the baseline Dynamic Programming logic) are implemented in \texttt{Python 3.11} and primarily executed on an AWS \texttt{c7i.24xlarge} instance.\footnote{https://aws.amazon.com/ec2/instance-types/} The instance provides 96 virtual CPUs and 192 GB of memory and is optimized for CPU compute. (No GPUs were required, though using GPUs may offer further efficiencies.) Computational experiments on other hardware types are included in Section \ref{sec:results}. Given the compute-intensive nature of state computation, the procedures \textsc{GSim} and \textsc{PSim} are compiled using the \texttt{numba 0.60} library for parallelization and faster execution. Neural networks for the PPO algorithm are implemented using \texttt{torch 2.1}. The PPO algorithm itself is custom-implemented, based on \cite{schulman_proximal_2017}. The environment in which the agents operate is custom built as an extension of the {\tt Env} environment in the {\tt gym} package from OpenAI (see \url{https://github.com/openai/gym}).

\subsection{Inference During Testing}
\label{sec:inference}

With our MetaRL model complete, we use a hand-designed set of 66 scenarios to test it, as described in Section \ref{sec:results} and Appendix \ref{66}. These scenarios fall both within and outside the training scenarios' distribution. During this testing, we only have access to the pre-trained actor models for GoalAgent and PortfolioAgent from the MetaRL algorithm. The state variable inputs are generated at each time step as described in Section \ref{sec:state}. The actor networks produce goal-taking and investment portfolio selection actions, which are implemented by the environment. As described in Appendix \ref{sec:curriculum}, we mitigate the risk of getting either unusually good or unusually bad results due to random initialization by training 5 models from scratch using identical scenarios in each epoch but different randomness for the 500 episodes within each epoch. The actions reported by RL inference using the MetaRL model are computed by sending the state space values to all 5 models within MetaRL and reporting the median of the 5 actions generated.

\section{Description Of The 66 Test Suite Cases}
\label{66}

Table \ref{tab:test_problems_stats} shows some descriptive statistics of our 66 test cases.
The specifics for the first 33 of the 66 test suite cases are given in Table \ref{tab:testcases_noinf}. These cases have no infusions. The specifics for the other 33 cases are given in Table \ref{tab:testcases_inf}. These are the same as the first 33 cases, except that they have infusions.
\begin{table}
\centering
\caption{\label{tab:test_problems_stats} %\small 
Descriptive statistics for the test suite of 66 GBWM problems. The suite comprises 66 new investor problems with varying horizons, initial wealth levels, number of goals, goal costs and utilities, and number of infusions.
}
%{\small
    \begin{tabular}{lHrrrrrrr}
\toprule
 & count & mean & std.~dev. & min & 25\% & 50\% & 75\% & max \\
\midrule
Time steps (in years) & 66 & 38 & 26 & 3 & 16 & 30 & 60 & 100 \\
Initial wealth & 66 & 93 & 23 & 12 & 100 & 100 & 100 & 126 \\
Number of goals & 66 & 16 & 23 & 1 & 1 & 4 & 10 & 60 \\
% Partial goal scenarios & 66 & 0 & 0 & 0 & 0 & 0 & 0 & 0 \\
Cost of all goals & 66 & 624,535 & 3,426,201 & 34 & 250 & 750 & 1830 & 20,000,000 \\
% Discounted goal cost & 66 & 301 & 702 & 8 & 74 & 133 & 220 & 4099 \\
Number of infusions & 66 & 9 & 19 & 0 & 0 & 0 & 1 & 99 \\
Total infusion amount & 66 & 29 & 80 & 0 & 0 & 1 & 21 & 588 \\
First infusion time & 66 & 5 & 11 & 0 & 0 & 0 & 1 & 49 \\
\bottomrule
\end{tabular}
%}
\end{table}

\begin{table}[]
    \centering
    \caption{Description of test cases without infusions.}
    \label{tab:testcases_noinf}
    \begin{tabular}{|p{1cm}|p{1cm}|p{1cm}|p{5.5cm}|p{4cm}|}
    \hline
    Case & $T$ & $W(0)$ & Non-zero goals $[t,C(t),U(t)]$ & Non-zero infusions $[t,I(t)]$ \\ \hline
    1 & 10 & 100 & $[10,150,1]$ & $-$ \\
    2 & 10 & 100 & $[10,200,1]$ & $-$ \\
    3 & 10 & 100 & $[10,400,1]$ & $-$ \\
    4 & 40 & 100 & $[40,600,1]$ & $-$ \\
    5 & 40 & 100 & $[40,1200,1]$ & $-$ \\
    6 & 40 & 100 & $[40,2400,1]$ & $-$ \\
    7 & 100 & 100 & $[100,50000,1]$ & $-$ \\
    8 & 100 & 100 & $[100,500000,1]$ & $-$ \\
    9 & 100 & 100 & $[100,20000000,1]$ & $-$ \\
    10 & 3 & 100 & $[2, 75, 0.9],[3, 75, 1] $ & $-$ \\
    11 & 20 & 100 & $[10, 200, 1.3],
[20, 500, 1] $ & $-$ \\
    12 & 20 & 100 & $[15, 200, 1.3],
[20, 300, 1] $ & $-$ \\
    13 & 35 & 100 & $[5, 50, 1],
[25, 500, .5],
[35, 1000, 1] $ & $-$ \\
    14 & 60 & 100 & $[15, 300, .7],
[30, 6000, 1.2]$, $
[45, 5000, .2],
[60, 20000, 1] $ & $-$ \\
    15 & 25 & 100 & $ [3, 30, .2],
[5, 70, .3],
[8, 70, .3]$, $
[25, 1000, 1]$ & $-$ \\
    16 & 40 & 100 & $[10, 150, 1.5],
[30, 400, 1]$, $
[35, 500, 1],
[40, 600, 1]$ & $-$ \\
17 & 20 & 100 & $[t,15,1]$ for $t\in\{2,4,6,\ldots,20\}$  & $-$ \\
18 & 20 & 100 & $[t,25,1]$ for $t\in\{2,4,6,\ldots,20\}$  & $-$ \\
19 & 20 & 100 & $[t,50,1]$ for $t\in\{2,4,6,\ldots,20\}$  & $-$ \\
20 & 20 & 100 & $[t,75,1]$ for $t\in\{2,4,6,\ldots,20\}$  & $-$ \\
21 & 60 & 100 & $[t,20,1]$ for $t\in\{1,2,3,\ldots,60\}$  & $-$ \\
22 & 60 & 100 & $[t,t,1]$ for $t\in\{1,2,3,\ldots,60\}$  & $-$ \\
23 & 60 & 100 & $[t,t,100+t]$ for $t\in\{1,2,3,\ldots,60\}$  & $-$ \\
24 & 60 & 100 & $[t,t,100-t]$ for $t\in\{1,2,3,\ldots,60\}$  & $-$ \\
25 & 60 & 100 & $[t,60-\frac{t}{2},1]$ for $t\in\{1,2,3,\ldots,60\}$  & $-$ \\
26 & 60 & 100 & $[t,60-\frac{t}{2},100+t]$ for $t\in\{1,2,\ldots,60\}$  & $-$ \\
27 & 60 & 100 & $[t,60-\frac{t}{2},100-t]$ for $t\in\{1,2,\ldots,60\}$  & $-$ \\
28 & 30 & 100 & $[3, 35, 1.5],
[6, 35, 1.3],
[9, 5, 0.4]$, $
[12, 50, 1],
[15, 15, 0.7],
[18, 5, 0.3]$, $
[21, 45, 0.6],
[24, 120, 0.9],
[27, 170, 1.1]$, $
[30, 160, 1]$ & $-$ \\
29 & 16 & 12 & $[16,34.25,26]$ & $-$ \\
30 & 16 & 21.63 & $[8,18.50,18], [16,34.25,26]$ & $-$ \\
31 & 16 & 38.99 & $[4,13.60,14], [8,18.50,18]$, $[12,25.18,22], [16,34.25,26]$ & $-$ \\
32 & 16 & 70.27 & $[2,11,66,12], [4,13.60,14]$, $[6,15.87,16], [8,18.50,18]$, $[10,21.59,20], [12,25.18,22]$, $[14,29.37,24], [16,34.25,26]$ & $-$ \\
33 & 16 & 126.67 & $[1,10.8,11], [2,11,66,12]$, $[3,12.60,13], [4,13.60,14]$, $[5,14.69,15], [6,15.87,16]$, $[7,17.14,17], [8,18.50,18]$, $[9,19.99,19], [10,21.59,20]$, $[11,23,32,21], [12,25.18,22]$, $[13,27.20,23], [14,29.37,24]$, $[15,31.72,25], [16,34.25,26]$ & $-$ \\
    \hline
    \end{tabular}
\end{table}

\begin{table}[]
    \centering
    \caption{Description of test cases with infusions. The base infusion amount for the odd-numbered test cases is $\Tilde{I}= \frac{W(0)}{10(T-1)}$, rounded up to the nearest integer. Aside from the infusions, these cases are identical to the 33 cases in Table \ref{tab:testcases_noinf}.}
    \label{tab:testcases_inf}
    \begin{tabular}{|p{0.9cm}|p{0.7cm}|p{0.9cm}|p{5.5cm}|p{4.5cm}|}
    \hline
    Case & $T$ & $W(0)$ & Non-zero goals $[t,C(t),U(t)]$ & Non-zero infusions $[t,I(t)]$ \\ \hline
    34 & 10 & 100 & $[10,150,1]$ & $[1,10]$ \\
    35 & 10 & 100 & $[10,200,1]$ & $[t,1.03^t\Tilde{I}]$ for $t\in\{1,..,T-1\}$ \\
    36 & 10 & 100 & $[10,400,1]$ & $[1,10]$ \\
    37 & 40 & 100 & $[40,600,1]$ & $[t,1.03^t\Tilde{I}]$ for $t\in\{1,..,T-1\}$ \\
    38 & 40 & 100 & $[40,1200,1]$ & $[6,12]$ \\
    39 & 40 & 100 & $[40,2400,1]$ & $[t,1.03^t\Tilde{I}]$ for $t\in\{1,..,T-1\}$ \\
    40 & 100 & 100 & $[100,50000,1]$ & $[21,19]$ \\
    41 & 100 & 100 & $[100,500000,1]$ & $[t,1.03^t\Tilde{I}]$ for $t\in\{1,..,T-1\}$ \\
    42 & 100 & 100 & $[100,20000000,1]$ & $[27,22]$ \\
    43 & 3 & 100 & $[2, 75, 0.9],[3, 75, 1] $ & $[t,1.03^t\Tilde{I}]$ for $t\in\{1,..,T-1\}$ \\
    44 & 20 & 100 & $[10, 200, 1.3],
[20, 500, 1] $ & $[6,12]$ \\
    45 & 20 & 100 & $[15, 200, 1.3],
[20, 300, 1] $ & $[t,1.03^t\Tilde{I}]$ for $t\in\{1,..,T-1\}$ \\
    46 & 35 & 100 & $[5, 50, 1],
[25, 500, .5],
[35, 1000, 1] $ & $[13,15]$ \\
    47 & 60 & 100 & $[15, 300, .7],
[30, 6000, 1.2]$, $
[45, 5000, .2],
[60, 20000, 1] $ & $[t,1.03^t\Tilde{I}]$ for $t\in\{1,..,T-1\}$ \\
    48 & 25 & 100 & $ [3, 30, .2],
[5, 70, .3],
[8, 70, .3]$, $
[25, 1000, 1]$ & $[11,14]$ \\
    49 & 40 & 100 & $[10, 150, 1.5],
[30, 400, 1]$, $
[35, 500, 1],
[40, 600, 1]$ & $[t,1.03^t\Tilde{I}]$ for $t\in\{1,..,T-1\}$ \\
50 & 20 & 100 & $[t,15,1]$ for $t\in\{2,4,6,\ldots,20\}$  & $[10,13]$ \\
51 & 20 & 100 & $[t,25,1]$ for $t\in\{2,4,6,\ldots,20\}$  & $[t,1.03^t\Tilde{I}]$ for $t\in\{1,..,T-1\}$ \\
52 & 20 & 100 & $[t,50,1]$ for $t\in\{2,4,6,\ldots,20\}$  & $[11,14]$ \\
53 & 20 & 100 & $[t,75,1]$ for $t\in\{2,4,6,\ldots,20\}$  & $[t,1.03^t\Tilde{I}]$ for $t\in\{1,..,T-1\}$ \\
54 & 60 & 100 & $[t,20,1]$ for $t\in\{1,2,3,\ldots,60\}$  & $[38,31]$ \\
55 & 60 & 100 & $[t,t,1]$ for $t\in\{1,2,3,\ldots,60\}$  & $[t,1.03^t\Tilde{I}]$ for $t\in\{1,..,T-1\}$ \\
56 & 60 & 100 & $[t,t,100+t]$ for $t\in\{1,2,3,\ldots,60\}$  & $[41,34]$ \\
57 & 60 & 100 & $[t,t,100-t]$ for $t\in\{1,2,3,\ldots,60\}$  & $[t,1.03^t\Tilde{I}]$ for $t\in\{1,..,T-1\}$ \\
58 & 60 & 100 & $[t,60-\frac{t}{2},1]$ for $t\in\{1,2,3,\ldots,60\}$  & $[45,38]$ \\
59 & 60 & 100 & $[t,60-\frac{t}{2},100+t]$ for $t\in\{1,2,\ldots,60\}$  & $[t,1.03^t\Tilde{I}]$ for $t\in\{1,..,T-1\}$ \\
60 & 60 & 100 & $[t,60-\frac{t}{2},100-t]$ for $t\in\{1,2,\ldots,60\}$  & $[49,43]$ \\
61 & 30 & 100 & $[3, 35, 1.5],
[6, 35, 1.3],
[9, 5, 0.4]$, $
[12, 50, 1],
[15, 15, 0.7],
[18, 5, 0.3]$, $
[21, 45, 0.6],
[24, 120, 0.9],
[27, 170, 1.1]$, $
[30, 160, 1]$ & $[t,1.03^t\Tilde{I}]$ for $t\in\{1,..,T-1\}$ \\
62 & 16 & 12 & $[16,34.25,26]$ & $[14,3]$ \\
63 & 16 & 21.63 & $[8,18.50,18], [16,34.25,26]$ & $[t,1.03^t\Tilde{I}]$ for $t\in\{1,..,T-1\}$ \\
64 & 16 & 38.99 & $[4,13.60,14], [8,18.50,18]$, $[12,25.18,22], [16,34.25,26]$ & $[15,6]$ \\
65 & 16 & 70.27 & $[2,11,66,12], [4,13.60,14]$, $[6,15.87,16], [8,18.50,18]$, $[10,21.59,20], [12,25.18,22]$, $[14,29.37,24], [16,34.25,26]$ & $[t,1.03^t\Tilde{I}]$ for $t\in\{1,..,T-1\}$ \\
66 & 16 & 126.67 & $[1,10.8,11], [2,11,66,12]$, $[3,12.60,13], [4,13.60,14]$, $[5,14.69,15], [6,15.87,16]$, $[7,17.14,17], [8,18.50,18]$, $[9,19.99,19], [10,21.59,20]$, $[11,23,32,21], [12,25.18,22]$, $[13,27.20,23], [14,29.37,24]$, $[15,31.72,25], [16,34.25,26]$ & $[16,21]$ \\
    \hline
    \end{tabular}
\end{table}

\section{Additional Comparisons of RL and DP Decisions} \label{sec:add_comp}

\begin{figure}[ht!]
\centering
DP: Investment Portfolio Decisions \hspace{.55in}RL Inference: Investment Portfolio Decisions\\
\includegraphics[scale=0.38]{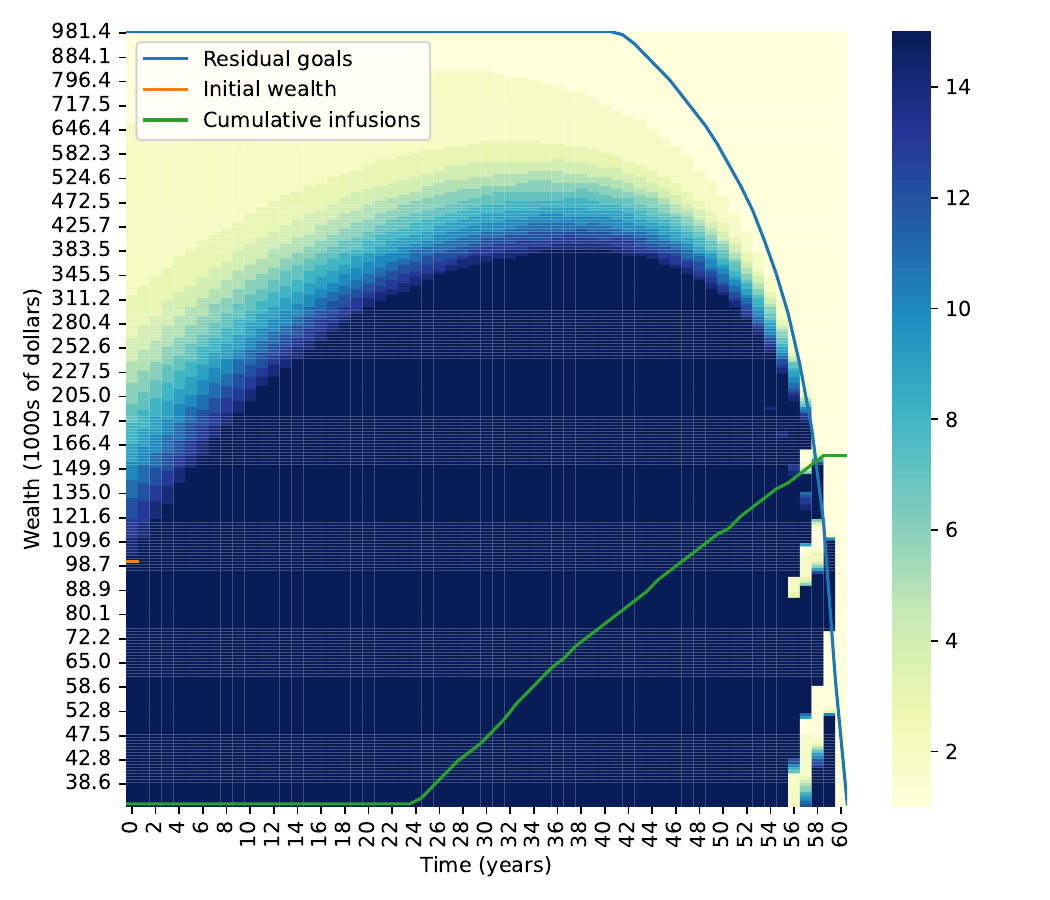}
\includegraphics[scale=0.38]{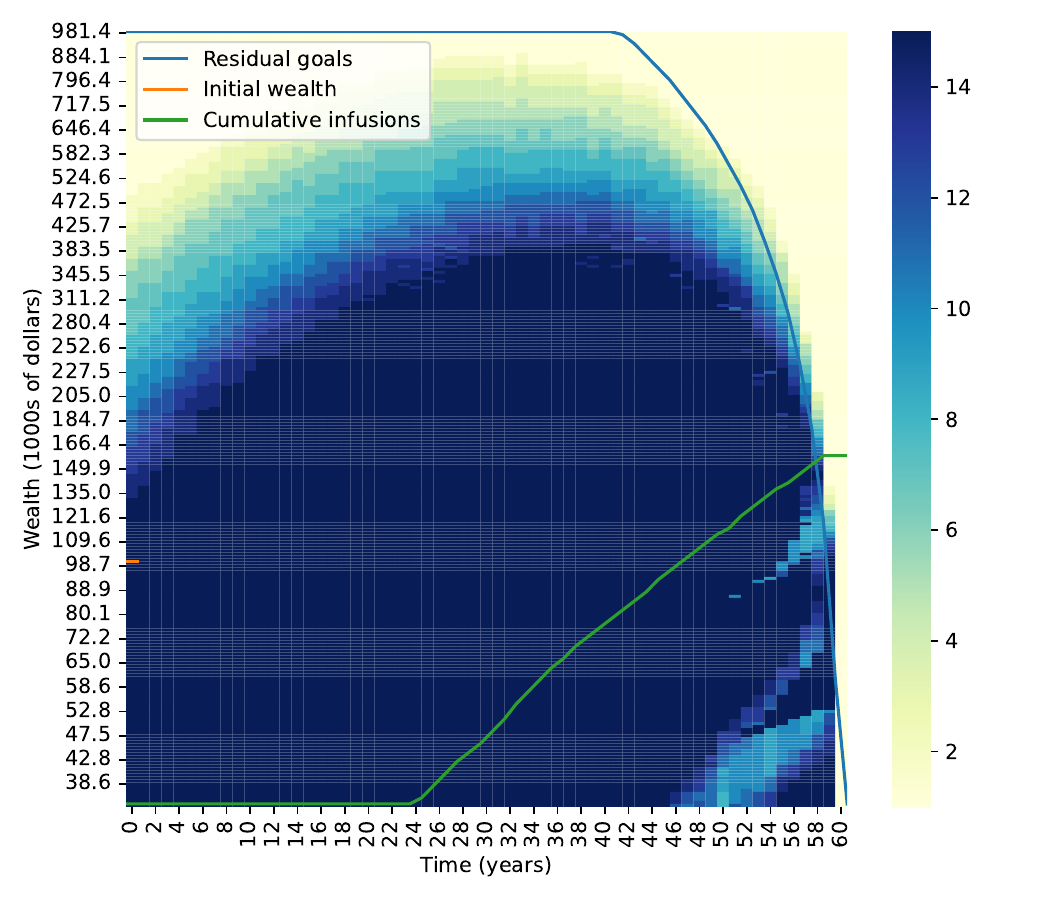}

DP: Goal-taking Decisions \hspace{1.05in}RL Inference: Goal-taking Decisions\\
\includegraphics[scale=0.38]{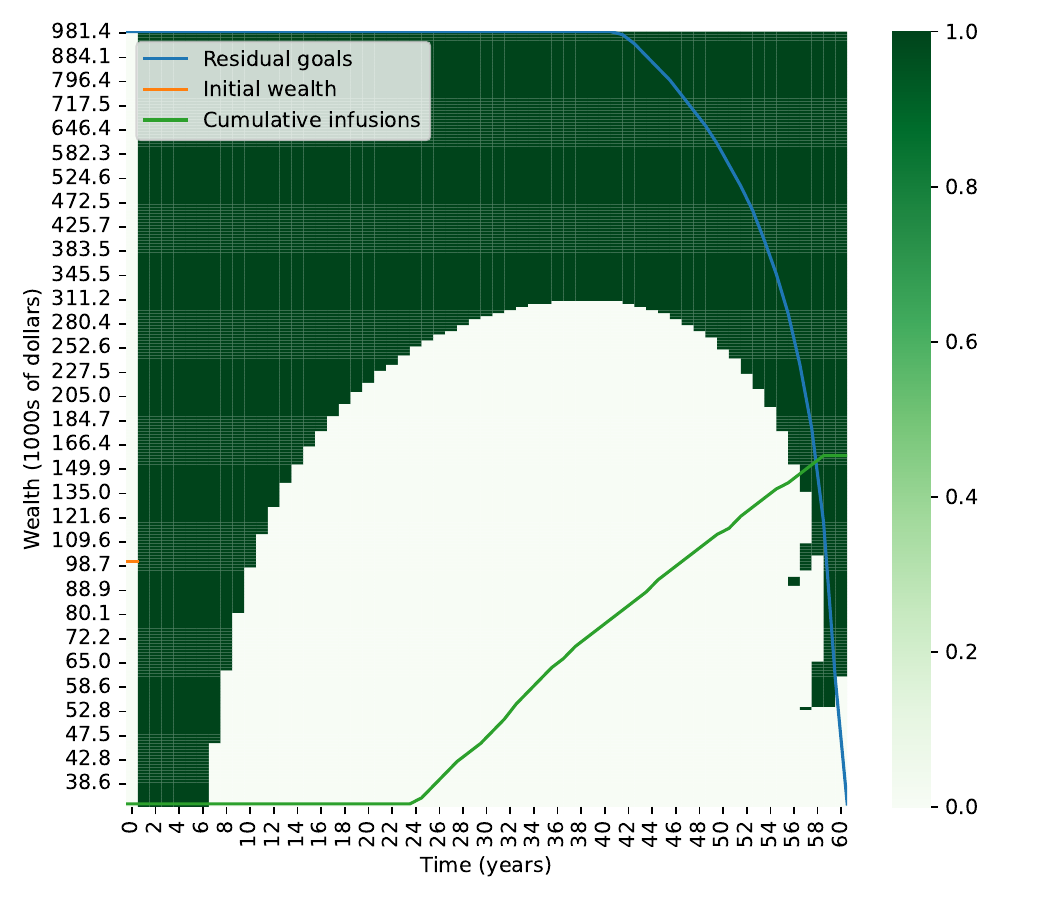}
\includegraphics[scale=0.38]{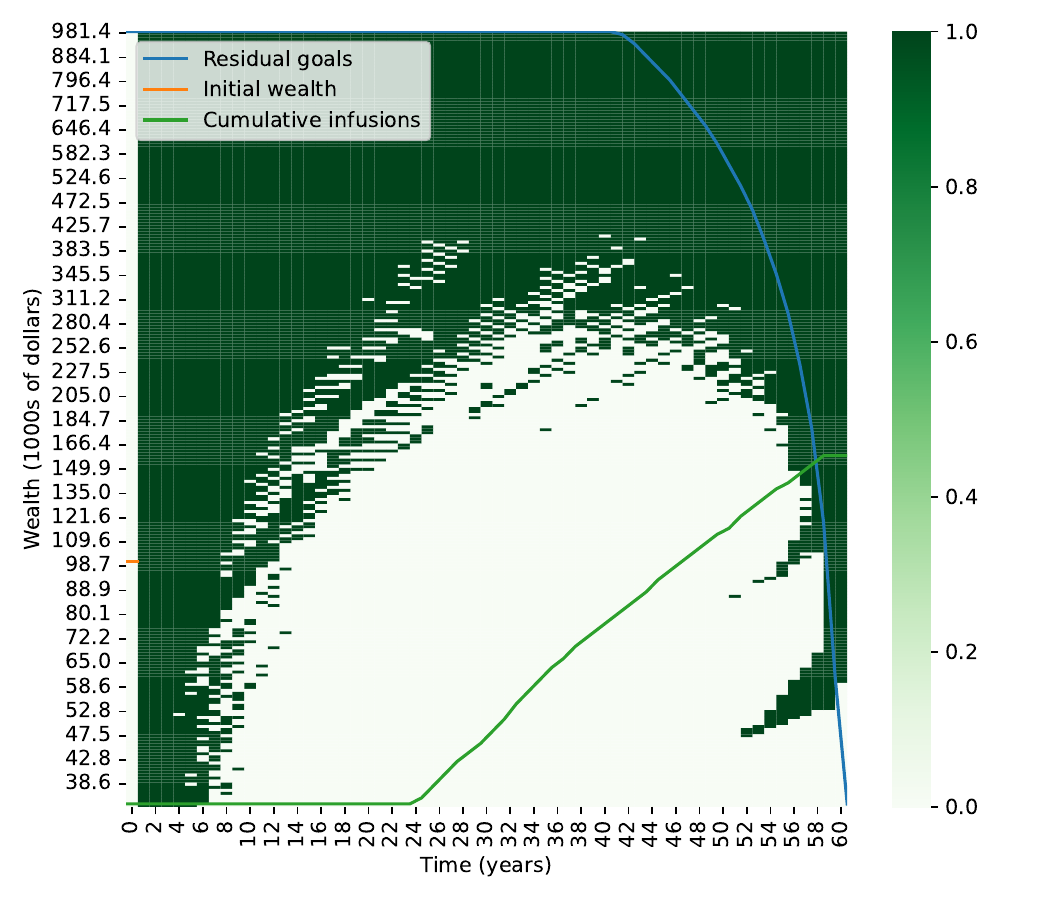}
\caption{\label{fig:policy_heat_maps_2} \small The same information shown in Figure \ref{fig:policy_heat_maps}, but for case 57 instead of case 20. Note that the green line representing cumulative infusions and the blue line representing the cost to fulfill the remaining goals are both truncated by the vertical axis's limited range.}
\end{figure}

The second representative example is case 57 of the 66 cases. From Table \ref{tab:testcases_inf}  in Appendix \ref{66}, we see that in this case: (1) There are 60 time steps (i.e., 60 years). (2) The initial wealth is 100 (thousand) dollars. (3) At each time step $t = 1,2,...,60$, there is a goal available that costs $t$ (thousand) dollars, which, if taken, creates a utility of $100-t$. (4) In each time step $t = 1,2,...,59$, there is an infusion worth $\frac{10}{59}1.03^t$ (thousand) dollars. A comparison of the decisions made by RL inference versus DP can be found in Figure \ref{fig:policy_heat_maps_2}.  In this case the optimal expected attained utility (i.e., the DP value function) is 3128 according to the DP backwards pass calculation. Using the method in Subsection \ref{sec:EUA}, our estimate for this value using the RL heatmaps will produce an average attained utility of 3090.

\section{Extending The Model To Concurrent Goals And/Or Partial Goals} \label{PCG}

The MetaRL model described in Appendix \ref{RLalgo} is designed to handle multiple goals, so long as there is at most one goal per time step. We now describe how to alter this model when we have multiple competing goals at the same time step and/or allow partial fulfillment of a goal (e.g., buying a less than ideal car, which generates less utility but also costs less than buying the ideal car).

Many of the updates to the concurrent and/or partial goals case are relatively straightforward. For example, for the computation of state inputs $W_{\min}(t)$, $W_{\max}(t)$, ${\bf{C}}_{\min}[t:]$, and ${\bf{C}}_{\max}[t:]$ from Table \ref{tab:observations}, we just replace $C(t)$, the cost of the single goal at time $t$, with the highest possible cost at time $t$, meaning the cost of taking all of the concurrent goals at time $t$ at their full (i.e., not partial) levels.  Similarly, to compute ${\bf{U}}_\mathrm{agg}[t:]$, we replace $U(t)$, the utility of the single goal at time $t$, with the highest possible utility at time $t$ from attaining all the concurrent goals at time $t$ at their full levels. 

This leaves the computation of $\gsim$ and $\psim$, which is a bit more complicated. In the concurrent/partial goals case, at each time $t$, there can be a number of possible costs and corresponding utilities corresponding to each possible combination of forgoing, partially fulfilling (in each available way), or fully fulfilling each of the concurrent goals. We form a pareto front of these possibilities by removing combinations that are dominated by any other combination; for example, we would remove a combination that generates a utility of 50 at a cost of \$40,000 if there is another combination that generates a utility of 52 at the same cost. In the pareto front of combinations that remain, the utility must increase as the cost increases. We will call the remaining goal combinations ${\bf G}(t)$, which contains the cost and corresponding utilities of these combinations in increasing order.  See \citet[]{das_dynamic_2022} for more details.

% The assumption behind having partial goals is that instead of having an all-or-nothing goal (spending cost $0$ or $C(t)$ at time $t$), one could have a sequence of possible cost levels at time $t$, each with its corresponding utility value. One may choose any one of these options. Similarly, concurrent goals imply that one may choose (at the same time $t$) to spend a certain amount towards one goal, or a different amount towards any number of other goals, or pick an arbitrary combination of goals to take. See \cite{das_dynamic_2022} for a detailed explanation of the problem. In order to support this functionality with the RL meta-model, we first convert the set of partial and concurrent goals into an ordered list ${\bf{G}}$ of costs and utilities as described in \cite{das_dynamic_2022}. This list is a pareto front of all possible combinations, implying that no additional utility may be achieved at the present time without spending additional money. 

% \begin{table}[h!]
% \centering
% \caption{\small Variables that constitute a complete description of a scenario. {\color{red} Needs more discussion in the text and caption.}}
% \label{tab:stateinputs2}
% {\small
% \begin{tabular}{|p{1.5cm}|p{2.3cm}|p{5.8cm}|}
% \hline
% Symbol & Dimension & Explanation \\ \hline
% ${\bf{C}}$ & $T$ (list(float)) & Cost of each partial goal at each time step \\
% ${\bf{U}}$ & $T$ (list(float)) & Utility of each partial goal at each time step \\
% \hline
% \end{tabular}
% }
% \end{table}

In the single all-or-nothing goal context of Appendix \ref{RLalgo}, $a_g(t) \in [0,1]$ is a measure of the confidence in taking ($a_g(t) \approx 1$) versus not taking ($a_g(t) \approx 0$) the goal at time $t$. In the current context, $a_g(t)$ is a measure of the confidence in taking the goal combination at time $t$ that corresponds to the highest approximated expected accumulated utility from time $t$ onwards versus the combination at time $t$ that corresponds to the second highest approximated expected accumulated utility from time $t$ onwards. The determination of the which two combinations at time $t$ have the two highest approximated expected accumulated utility from time $t$ onwards comes from our computation for determining $\gsim$.

To compute $\gsim$, we first extend the computation of the $P$-vectors ${\bf E}_\mathrm{take}$ and ${\bf E}_\mathrm{skip}$ in Algorithm \ref{alg:gsim} to all the combinations in ${\bf G}(t)$. This results in a matrix ${\bf{E}}$ of dimension $P\times |{\bf{G}}(t)|$, where $|{\bf{G}}(t)|$ is the number of goal combinations in ${\bf G}(t)$. Each column in ${\bf E}$ is determined by following Algorithm \ref{alg:gsim} for computing ${\bf E}_\mathrm{take}$, but forcing taking the goal combination corresponding to the column to be the first goal taken (i.e., the first entry in ${\bf sortU}$). 
The highest approximated expected utility for every goal combination is given by taking the maximum of each column of ${\bf{E}}$, and the logistic in \textsc{GSim} is computed between the highest and second-highest values of the column-maxima. The goal combinations that correspond to these two columns are the two goal combinations that $a_g(t)$ chooses between.
Note that our approach here conforms to Appendix \ref{RLalgo}, for which $|{\bf{G}}|=2$, corresponding to not taking or taking the single all-or-nothing goal. 

For computing $\psim$, we start with the same matrix ${\bf{E}}$ calculated for $\gsim$. However, instead of computing column maxima, we determine the index of the row that contains the largest component in ${\bf{E}}$. This index approximates the best portfolio $p$. As in Algorithm \ref{alg:psim}, we define $\psim$ to be this index divided by $P-1$ to normalize $\psim$ to be between 0 and 1.

\section{Extending The Model To Stochastic Inflation}
\label{sec:app_stoch_infln}

In this section we show how our MetaRL model is extended to cases where inflation is a stochastic process. We assume that the evolution of $I_{\mathrm{infl}}(t)$, the instantaneous rate of inflation over time $t$, is governed by the \cite{vasicek_equilibrium_1977} model, whose dynamics are governed by the stochastic differential equation
\begin{equation}\label{SDE1}
dI_{\mathrm{infl}} = -\kappa_{\mathrm{infl}}(I_{\mathrm{infl}} - \theta_{\mathrm{infl}})dt+\sigma_{\mathrm{infl}} dW,
\end{equation}
where $W$ is a Wiener process. The Vasicek model is an Ornstein–Uhlenbeck process, where $\theta_{\mathrm{infl}}$ is the mean value of $I_{\mathrm{infl}}$, $\kappa_{\mathrm{infl}}$ is the constant strength of the reversion to this mean value, and $\sigma_{\mathrm{infl}}$ is a volatility constant representing the strength of the randomness. This process admits negative values for inflation and hence, deflationary environments are possible in this model. 
The solution to equation (\ref{SDE1}) between times $t$ and $\tau$ is 
\begin{equation}\label{inflsoln}
I_{\mathrm{infl}}(\tau) = I_{\mathrm{infl}}(t)e^{-\kappa_{\mathrm{infl}}(\tau-t)} + \theta_{\mathrm{infl}}(1-e^{-\kappa_{\mathrm{infl}}(\tau-t)}) + \sigma_{\mathrm{infl}}\sqrt{\frac{1-e^{-2\kappa_{\mathrm{infl}}(\tau-t)}}{2\kappa_{\mathrm{infl}}}} Z,
\end{equation}
where $Z$ is a standard normal random variable. During training for the MetaRL model, we use this equation to update the inflation from the previous year to the current year by replacing $t$ with $t-1$ and substituting $\tau =t$ in equation \eqref{inflsoln}. 

We must multiply each future real infusion and each real goal cost by the cumulative effect of the inflation. We split this cumulative effect into the effect of the inflation between time 0 and the current time $t$, and the effect from time $t$ to the future time $\tau$ when the infusion occurs or the goal is available. To calculate the cumulative effect between time 0 to the current time $t$, we simply use the inflation that has been generated using equation \eqref{inflsoln} up to the current time $t$: 
$$
I_{\mathrm{cum,past}}(t) = 
e^{\int_0^t I_{\mathrm{infl}}(s)ds} =
e^{I_{\mathrm{infl}}(0) + I_{\mathrm{infl}}(1) +\cdots+ 
I_{\mathrm{infl}}(t-1)}.
$$
To account for the future effect of inflation, we compute $\bar{I}_{\mathrm{cum}}(\tau)$, the expected value of the cumulative effect of inflation between times $t$ and $\tau$, which is
\begin{equation}
\bar{I}_{\mathrm{cum}}(\tau) = E\left[e^{\int_t^\tau I_{\mathrm{infl}}(s)ds} \right].    
\end{equation}
We then adjust the real infusion or real goal cost by multiplying the time 0 cost by $I_{\mathrm{cum,past}}(t)$ and then by $\bar{I}_{\mathrm{cum}}(\tau)$.
The explicit formula for $\bar{I}_{\mathrm{cum}}(\tau)$ is known (see, for instance, \cite{vasicek_equilibrium_1977}) to be
\begin{equation}
\bar{I}_{\mathrm{cum}}(\tau)
= e^{I_{\mathrm{infl}}(t)A(\tau)- B(\tau) },
\end{equation}
where
\begin{eqnarray}
A(\tau) &=& \frac{1}{\kappa_{\mathrm{infl}}}\big(1-e^{-\kappa_{\mathrm{infl}}(\tau-t)}\big) \mbox{ and}   \\
B(\tau) &=& \left(\theta_{\mathrm{infl}} - \frac{\sigma_{\mathrm{infl}}^2}{2\kappa_{\mathrm{infl}}^2} \right)\big(A(\tau)-(\tau-t)\big) - \frac{\sigma_{\mathrm{infl}}^2}{4\kappa_{\mathrm{infl}}} A^2(\tau).   
\end{eqnarray}
% As an example of using this formula, if the cost of a goal at time $=\tau$ (years) adjusts with inflation and its time $=t$ cost is \$120,000, then \$120,000$\times \bar{I}_{\mathrm{cum}}(\tau)$ is the expected value at time $=t$ of the goal's nominal cost at time $=\tau$.

To account for the stochastic inflation in the MetaRL model, we make three straightforward changes. The first change is to introduce inflation into the model, which is generated via the Vasicek process in equation \eqref{inflsoln}. This is used to update all real infusions and real costs to the current time $t$. The second change is to introduce the new state variable $I_\mathrm{norm}$, which is the current instantaneous inflation, $I_{\rm infl}(t)$, normalized by the mean inflation in the Vasicek process, $\theta_{\mathrm{infl}}$:
$$
I_\mathrm{norm} = \frac{I_{\rm infl}(t)}{\theta_{\mathrm{infl}}}.
$$
We add $I_\mathrm{norm}$ to the list of the other $3K+5$ state variables in Table \ref{tab:observations}. The calculations for the state variables $t_{\mathrm{norm}}$ and ${\bf{U}}_{\mathrm{agg}}[t:]$ remain the same.
The third change is to the calculations of the remaining state variables, $W_{\min}(t)$, $W_{\max}(t)$, ${\bf{C}}_{\min}[t:]$, ${\bf{C}}_{\max}[t:]$, $\gsim$, and $\psim$ from Table \ref{tab:observations}. In the calculations of these $2K+4$ state variables, we simply replace \textsc{DiscountSum} and \textsc{DiscountVec} defined in Algorithm \ref{alg:discount} with \textsc{DiscountSumInfl} and \textsc{DiscountVecInfl} defined in Algorithm \ref{alg:inflatediscount}. The only difference between Algorithm \ref{alg:discount} and Algorithm \ref{alg:inflatediscount} is that Algorithm \ref{alg:inflatediscount} takes the average projected cumulative inflation, $\bar{I}_{\mathrm{cum}}$, into account, so future infusions and costs are adjusted for projected average future inflation.

\begin{algorithm}[H]
{\footnotesize
\caption{Discounting future goal costs with stochastic inflation}
\label{alg:inflatediscount}
\begin{algorithmic}
\Procedure{DiscountVecInfl}{${\bf{C}}[t:]$, $p$, $z$, $I_{\mathrm{curr}}$, $\theta_{\mathrm{infl}}$, $\kappa_{\mathrm{infl}}$, $\sigma_{\mathrm{infl}}$}
    \State ${\bf{C}}_\mathrm{disc} \gets$ initialize vector of zeros of same length as ${\bf{C}}[t:]$, which is $T-t+1$
    \For{$\tau$ in $[0:$ length(${\bf{C}}[t:]$)}
        \State $A \gets \frac{1}{\kappa_{\mathrm{infl}}}\big(1-e^{-\kappa_{\mathrm{infl}}\tau}\big)$
        \State $B \gets \left(\theta_{\mathrm{infl}} - \frac{\sigma_{\mathrm{infl}}^2}{2\kappa_{\mathrm{infl}}^2} \right)\big(A-\tau \big) - \frac{\sigma_{\mathrm{infl}}^2}{4\kappa_{\mathrm{infl}}} A^2$
        \State $I_{\mathrm{cum,avg}} \gets \exp\left(I_{\mathrm{curr}}A - B \right)$
        \State ${{C}}_\mathrm{disc}(\tau) \gets {{C}}(\tau)\cdot\mathrm{exp}\left[ {-(\mu_p -\frac{1}{2}\sigma_p^2)h\tau - \sigma_p  z \sqrt{h\tau}}\right] \cdot
        I_{\mathrm{cum,avg}}$ 
    \EndFor
    \State \textbf{return} ${\bf{C}}_\mathrm{disc}$
\EndProcedure
\\
\Procedure{DiscountSumInfl}{${\bf{C}}[t:]$, $p$, $z$, $I_{\mathrm{curr}}$, $\theta_{\mathrm{infl}}$, $\kappa_{\mathrm{infl}}$, $\sigma_{\mathrm{infl}}$}
    \State \textbf{return} sum$\big(\textsc{DiscountVecInfl}({\bf{C}}[t:]$, $p$, $z$, $I_{\mathrm{curr}}$, $\theta_{\mathrm{infl}}$, $\kappa_{\mathrm{infl}}$, $\sigma_{\mathrm{infl}}$) \big)
\EndProcedure
\end{algorithmic}
}
\end{algorithm}

\end{document}